\documentclass[letterpaper, 10 pt, conference]{ieeeconf}  % Comment this line out if you need a4paper
\IEEEoverridecommandlockouts
\usepackage{cite}
\usepackage{amsmath,amssymb,amsfonts}
\usepackage{algorithmic}
\usepackage{graphicx}
\usepackage{textcomp}
\usepackage{subcaption}
\usepackage{xcolor}

\usepackage{hyperref}
\def\BibTeX{{\rm B\kern-.05em{\sc i\kern-.025em b}\kern-.08em
    T\kern-.1667em\lower.7ex\hbox{E}\kern-.125emX}}

\usepackage{lipsum}
\usepackage{tikz}
\usetikzlibrary{positioning}
\usetikzlibrary{calc, external, fit}
\usepackage{multirow, makecell}
\usepackage{hhline}
\usepackage{array, booktabs}
\usepackage{siunitx}
\usepackage{float} 
\usepackage{lipsum}
\usepackage{fancyhdr,graphicx,amsmath,amssymb}
\usepackage[ruled,vlined]{algorithm2e}
\usepackage{url}
\newcommand{\IBRLn}{DRLR}

\title{Solving Robotics Tasks with Prior Demonstration via Exploration-Efficient Deep Reinforcement Learning}
\author{Chengyandan Shen$^{1}$ and Christoffer Sloth$^{2}$
\thanks{This work was supported by Unicontrol, funded by Innovation Fund Denmark, grant number 1044-00117B.}
\thanks{$^{1}$Chengyandan Shen is with Unicontrol, Odense, Denmark. {\tt\small chengyandan.shen@unicontrol.io}}
\thanks{$^{2}$Christoffer Sloth is with the Maersk McKinney Møller Institute, University of Southern Denmark, Odense, Denmark {\tt\small chsl@mmmi.sdu.dk}}
}

\begin{document}

\maketitle
\begin{abstract}
\noindent
This paper proposes an exploration-efficient Deep Reinforcement Learning with Reference policy (\IBRLn) framework for learning robotics tasks that incorporates demonstrations. \IBRLn{} framework is developed based on an algorithm called Imitation Bootstrapped Reinforcement Learning (IBRL). We propose to improve IBRL by modifying the action selection module. The proposed action selection module provides a calibrated Q-value, which mitigates the bootstrapping error that otherwise leads to inefficient exploration. Furthermore, to prevent the RL policy from converging to a sub-optimal policy, SAC is used as the RL policy instead of TD3. The effectiveness of our method in mitigating bootstrapping error and preventing overfitting is empirically validated by learning two robotics tasks: bucket loading and open drawer, which require extensive interactions with the environment. Simulation results also demonstrate the robustness of the \IBRLn{} framework across tasks with both low and high state-action dimensions, and varying demonstration qualities.
To evaluate the developed framework on a real-world industrial robotics task, the bucket loading task is deployed on a real wheel loader. The sim2real results validate the successful deployment of the \IBRLn{} framework.
\end{abstract}
\section{Introduction}
\noindent
Model-free Deep Reinforcement Learning (DRL) has shown great potential in learning continuous control tasks in robotics \cite{allshire2022transferring, qi2023hand, rudin2022learning, haarnoja2018learning, nguyen2019review, ibarz2021train}. 
However, there are still challenges that limit the widespread applicability of these methods in real-world robotic applications. One major challenge is the poor sample efficiency of learning with model-free DRL, even relatively simple tasks can require millions of interaction steps, while learning policies from high-dimensional observations or complex environments may require significantly more interactions \cite{haarnoja2018soft, kaiser2019model, raffin2021stable}. A primary cause for the poor sample efficiency is on-policy learning \cite{haarnoja2018soft}, since some of the most widely used DRL algorithms, such as A3C \cite{mnih2016asynchronous} and PPO \cite{schulman2017proximal}, require new interactions with the environments for each gradient step.
Consequently, on-policy DRL is often impractical for real-world systems, as allowing untrained policies to interact with real systems can be both costly and dangerous.
Even when learning occurs solely in simulation, it is still preferred to utilize previously collected data instead of starting from scratch \cite{levine2020offline}. 
On the other hand, off‑policy DRL methods improve sample efficiency by reusing past experience, and have demonstrated strong performance on continuous control tasks \cite{bengio2009continuous,fujimoto2018addressing,haarnoja2018soft,fujimoto2019off,fujimoto2021minimalist,kumar2020conservative}. However, for complex robotics tasks where data collection itself is expensive, e.g., in construction machines, educational agents, or medical devices, even off‑policy approaches become costly when the DRL policy requires extensive explorations. In these scenarios, improving exploration efficiency is as crucial as sample efficiency to reduce the exploration needed for achieving a good policy.

\begin{figure}[t]
    \centering
    \includegraphics[width=8.5cm]{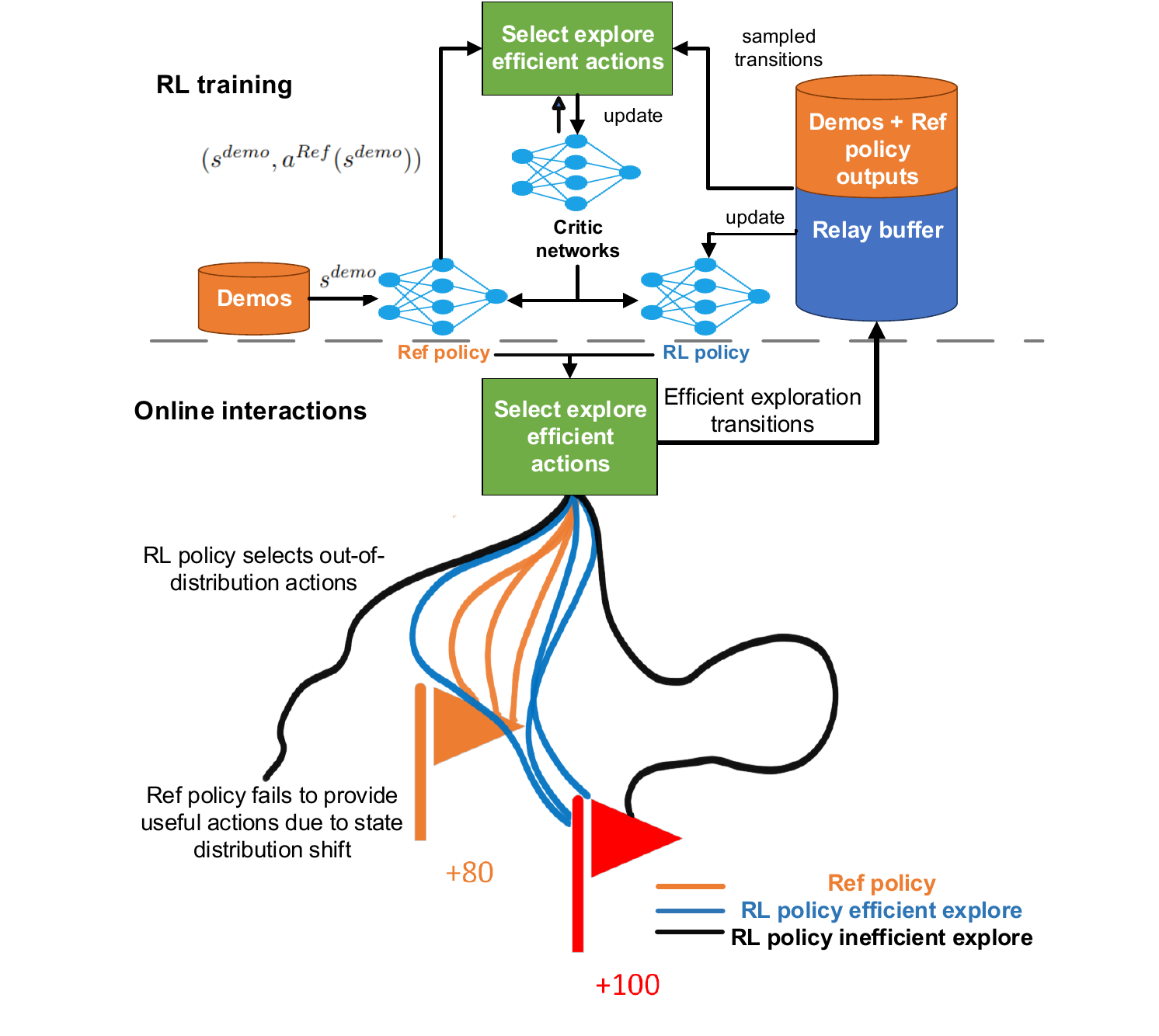 }
    \caption{Overview of the proposed exploration-efficient \IBRLn{} framework. The proposed framework extends a sample-efficient DRL-Ref method with a simple action selection module to mitigate inefficient explorations caused by (1) Bootstrapping error leads to the RL policy selecting out-of-distribution actions. (2) Ref policy fails to provide good actions under state distribution shifts. }
    \label{fig:general_framework}
\end{figure}

Thus, effectively leveraging prior demonstrations to facilitate efficient exploration is considered a promising strategy for the broad application of off‑policy DRL in real‑world industrial robotics. Two main research directions have emerged to achieve this goal:

\textbf{Offline-to-online DRL:}
Pretraining the DRL with prior expert demonstrations, and continuing training with online data, has shown its impressive performance in exploration efficiency \cite{vecerik2017leveraging,nair2020awac,uchendu2023jump,zhou2024efficient,goecks2019integrating, lee2022offline}. Early studies start training by mixing offline demonstrations and online interaction in the replay buffer, and use a prioritized replay mechanism to enable the RL policy to efficiently explore \cite{vecerik2017leveraging,song2022hybrid}. More recent approaches separate offline pretraining from online fine‑tuning and report superior exploration efficiency \cite{gao2018reinforcement, goecks2019integrating, nair2020awac,lee2022offline}. In offline training, a Behavior Cloning (BC) loss or KL divergence is typically employed to encourage the RL policy to closely follow the behavior policy, which is used to generate the demonstrations, thereby facilitating efficient exploration in online interactions. However, when transferring to the online interacting phase, some methods are required to “recalibrate” the offline Q-estimates to the new online distribution to keep the learning stable and mitigate forgetting of pre-trained initializations \cite{nair2020awac, uchendu2023jump, ball2023efficient}.
\\

\textbf{DRL-Ref policy:}
Some novel studies have proposed to explicitly integrate a reference policy, trained from the prior demonstration to guide DRL training \cite{zhang2023policy, IBRL}. In these works, a standalone reference policy is trained using the offline demonstration and then used to provide additional guidance in the DRL online learning phase. In this work, we consider the Imitation Bootstrapped Reinforcement Learning (IBRL) framework as an ideal approach for learning robotics tasks with prior demonstrations, as it avoids catastrophic forgetting of pre-trained initializations and automatically balances offline and online training \cite{IBRL}. 

However, the IBRL framework is built on off-policy RL and Imitation Learning (IL). It risks the same challenges brought by bootstrapping error in off-policy RL \cite{kumar2019stabilizing, kumar2020conservative, fujimoto2019off, fujimoto2021minimalist}, where the target critic and actor networks are updated using out-of-distribution (OOD) actions with overestimated Q-value \cite{kumar2019stabilizing, kumar2020conservative}. Meanwhile, the IL policy in IBRL could also face the state distribution shift \cite{hussein2017imitation}, when OOD actions keep getting selected.
To tackle these challenges, in this work, we propose an exploration-efficient DRL with Reference policy (\IBRLn{}) framework, shown in Fig.~\ref{fig:general_framework}, and make the following contributions:
\begin{enumerate}
    \item Identify and analyze the main cause of the failure cases trained with the IBRL framework: Distribution shift due to bootstrapping error.
    \item Propose a simple action selection module and employ a Maximum Entropy RL to mitigate inefficient explorations caused by bootstrapping error and convergence on a sub-optimal policy due to overfitting.
    \item Demonstrate the effectiveness and robustness of the proposed framework on tasks with both low and high state-action dimensions, and demonstrations of different quality.
   \item Showcase an implementation and deployment of the proposed framework on a real industrial task.
\end{enumerate}

\section{Problem statement}
\begin{figure}[htb]
  \centering
  \begin{subfigure}[b]{0.45\columnwidth}
    \centering
    \includegraphics[width=\linewidth]{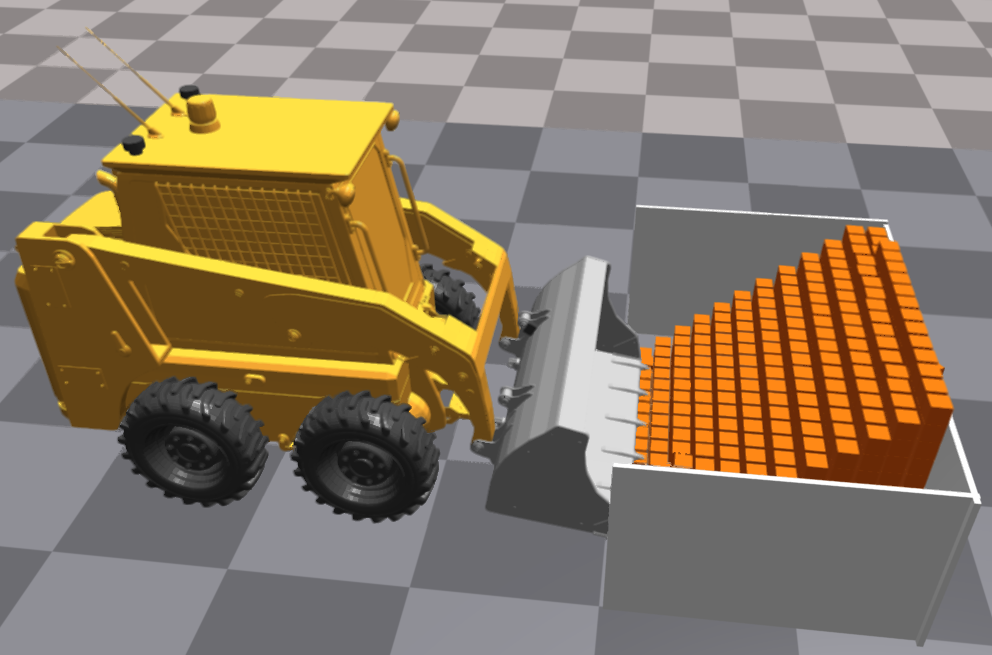}
    \caption{Bucket loading}
    \label{fig:1}
  \end{subfigure}
  \hfill
  \begin{subfigure}[b]{0.48\columnwidth}
    \centering
    \includegraphics[width=\linewidth]{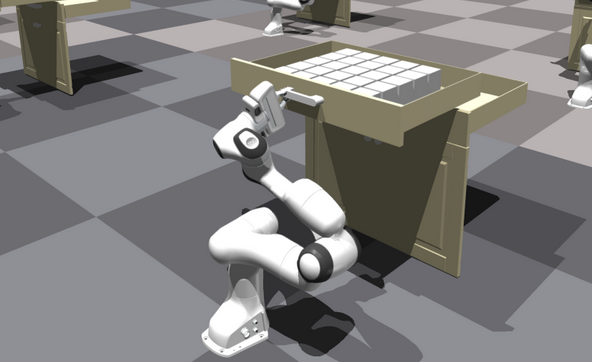}
    \caption{Open drawer}
    \label{fig:3}
  \end{subfigure}
  \caption{Selected tasks for testing the proposed framework.}
  \label{fig:examples}
\end{figure}

% In this work, we proposed an exploration-efficient framework, shown in Fig.~\ref{fig:general_framework}. 
The proposed framework is generalized towards learning robotics tasks with the following problems: 1) Collecting a large amount of data is costly. 2) Learning requires extensive interactions. 3) A small number of expert demonstrations are available.
Based on the characteristics, bucket loading \cite{Shen2024} and open drawer \cite{cab_open} tasks are selected to evaluate the effectiveness of the proposed framework. The task environments are shown in Fig.~\ref{fig:examples}.

Compared to the selected DRL-Ref framework, IBRL, the proposed framework attempts to mitigate distribution shift caused by bootstrapping errors and prevent convergence to a suboptimal policy from overfitting to the demonstrations.

\textbf{Bootstrapping error} can arise in off-policy RL when the value function is updated using Bellman backups. It occurs because the target value function and policy are updated using OOD actions with overestimated Q-values \cite{kumar2019stabilizing}.
Studies have shown that bootstrapping error can lead to unstable training and even divergence from the optimal policy \cite{kumar2019stabilizing,kumar2020conservative}, especially when the current policy output is far from the behavior policy, which is used to generate the transitions in the replay buffer \cite{fujimoto2019off, fujimoto2021minimalist, kumar2020conservative, kumar2019stabilizing}. 
% This is because during value networks updating using Bellman backup:
% \begin{align}
%     L\left(\theta_{Q}\right)=\mathbb{E}_{s_t, a_t, r_{t}, s_{t+1} \sim \beta, r_{t} \sim E}\left[\left(\hat{Q}(s_t, {a}_t)-y_{t}\right)^{2}\right]
% \end{align}
% where
% \begin{align}
%     y_{t}=r_{t}+\gamma Q_{max}\left(s_{t+1}, \pi_{\phi}(s_{t+1}) \mid \theta'_{Q}\right)
% \end{align}
% The Q-value $\hat{Q}(s_t, {a}_t)$ is estimated with actions sampled from the behavior policy that produced the dataset in the replay buffer, while the target value $y_t$ is estimated using current policy $\pi_{\phi}(s_{t+1})$. If the output of the current policy does not correspond with the distribution under the offline dataset, the target value $y_t$ may be a poor estimate. Furthermore, the target value $y_t$ is calculated by maximizing the $Q(s_{t+1}, \pi_{\phi}(s_{t+1}))$. $Q(s_{t+1}, \pi_{\phi}(s_{t+1}))$ is only reliable when the state-action pair is from the same distribution as the training set \cite{kumar2019stabilizing}. However, $\pi_{\phi}(s_{t+1})$ could propose OOD actions with 'fake' high Q-values. Since $\pi_{\phi}$ is updated by maximizing Q-value $Q(s_{t}, \pi_{\phi}(s_{t}))$, the agent tends to prefer these OOD actions, resulting in bootstrapping error.

In the IBRL, the critic (value) functions's parameters $\phi$ are updated with the following Bellman backup \cite{IBRL}:
\begin{align}
    L\left(\phi\right)=\mathbb{E}_{s_t, a_t, r_{t}, s_{t+1} \sim \mathcal{B}}\left[\left(\hat{Q}_{\phi}(s_t, {a}_t)-Q_{\phi}\right)^{2}\right]
    \label{eqa:IBRL_crtic_update}
\end{align}
where
\begin{align}
    Q_{\phi} \leftarrow r_{t}+\gamma \underset{a^{\prime} \in\left\{a_{t+1}^{\mathrm{IL}}, a_{t+1}^{\mathrm{RL}}\right\}}{\operatorname{argmax}} Q_{\phi'}\left(s_{t+1}, a^{\prime}\right)
    \label{eqa.target}
\end{align}

and $\hat{Q}_{\phi}(s_t, {a}_t)$ is the estimated Q-value with states and actions sampled from the replay buffer $\mathcal{B}$, while the target value $Q_{\phi'}\left(s_{t+1}, a^{\prime}\right)$ in \eqref{eqa.target} is estimated using the current RL policy $a_{t+1}^{\mathrm{RL}}$ or IL policy $a_{t+1}^{\mathrm{IL}}$. 
IBRL training starts with a replay buffer mixed with expert demonstrations and transitions collected during interactions, which introduces a mismatch between the current RL policy and behavior policy. Although IBRL allows for selecting actions from the IL policy, whose output is closer to the behavior policy in the demonstration, it relies on an accurate value estimation between $Q_{\phi'}\left(s_{t+1}, a_{t+1}^{\mathrm{RL}}\right)$ and $Q_{\phi'}\left(s_{t+1}, a_{t+1}^{\mathrm{IL}}\right)$. However, because of the exploration noises during the online interaction, the future rollout states $s_{t+1}$ sampled from $\mathcal{B}$ are likely OOD relative to the offline demo buffer, $\mathcal{D}$ \cite{lee2022offline, nakamoto2023cal, zhou2024efficient}. When the IL policy proposes actions in these OOD states, the critic networks have no prior data for these state-action pairs and could assign a lower Q-value compared to the OOD actions proposed by the RL agent. As a result, the lower bounds brought by the IL policy fail if the RL policy is updated with bad OOD actions with an overestimated Q-value.
Such errors could be corrected by attempting the OOD action in the online interaction and observing its actual Q-value, but in turn, bringing insufficient policy exploration. Thus, finding a reliable and calibrated Q-value estimation is crucial for mitigating the bootstrapping error \cite{nakamoto2023cal}. 

Another disadvantage of bootstrapping error is that OOD actions selected by the RL policy during online interaction can lead to 
\textbf{state distribution shift}. When the IL agent fails to provide high-quality actions for the unseen interaction states, the exploration efficiency of IBRL will be degraded.  
Furthermore, although the IBRL has stated that both Twin Delayed DDPG (TD3) and SAC can be employed as RL policies for continuous control tasks \cite{IBRL}, the authors exclusively used TD3 in their experiments due to its strong performance and high sample efficiency in challenging image-based RL settings.
However, we argue that the deterministic RL algorithm, TD3, is less suitable for high-dimensional, continuous state-based tasks, as it is more prone to overfitting offline data, converging to suboptimal policies, and suffering from inefficient exploration \cite{haarnoja2018soft}. 
To prevent the RL policy from convergence to a suboptimal policy because of \textbf{overfitting}. A maximum Entropy, stochastic RL, Soft Actor-Critic (SAC), is considered.

% The task can be mainly divided into two phases: 1. Approaching the environment. 2. Interaction with the environment. The approach phase is less critical to achieve the task, and exploration during this stage is unlikely to yield significant rewards, though it remains necessary to complete the task. While the interaction phase is crucial for task success, exploration at this stage has the potential to yield higher rewards.

% The proposed framework is developed based on Imitation Bootstrapped Reinforcement Learning (IBRL) \cite{IBRL}. IBRL introduces a novel RL framework with prior demonstrations, integrating both IL and RL agents during training. The IL agent provides an alternative action to facilitate RL agent exploration while preventing oversampling of demonstrations, and eliminating the need for IL loss regulation. 
% For tasks with limited demonstrations, the key is to get the most information from them. A simple IL, such as Behavior Cloning (BC) using Deep Neural Networks (DNN), only mimics the actions in the demonstrations given the same states. However, Bayesian Neural Networks (BNN), which train their network weights as Gaussian distributions, can offer additional uncertainty estimation for the predicted actions \cite{kononenko1989bayesian}. This study explores how effectively a BC agent, trained with BNN, can be integrated into the IBRL framework to actively sample actions with higher uncertainties, thereby improving the efficiency of action exploration and training for the RL agent.
\label{problem_state}

\section{Preliminaries}
This section presents an overview of Maximum Entropy RL and IBRL.

\subsection{Maximum Entropy Deep Reinforcement Learning}
For sample efficiency, off-policy DRL methods have been widely studied due to their ability to learn from past experiences. However, studies have also found that the off-policy DRL method struggles to maintain stability and convergence in high-dimensional continuous state-action spaces \cite{haarnoja2018soft}. To tackle this challenge, maximum entropy DRL has been proposed. 

As the state-action spaces are continuous in the selected robotics tasks, we consider a Markov Decision Processes (MDP) with continuous state-action spaces: An agent explores and interacts with an environment, at each time step $t$, the agent observes the state $s_t$, takes action $a_t$ based on RL policy $\pi_{\theta}$ with parameters $\theta$, and receives rewards $r_t$.
Different from standard RL, which aims to find a policy that maximizes the expected return:
\begin{equation}
J(\pi) = \sum_{t=0}^{T} \mathbb{E}_{(s_t, a_t) \sim \theta_{\pi}} \gamma^t r(s_t, a_t), \quad \gamma \in (0, 1] 
\end{equation}
maximum entropy DRL aims to maximize the discounted reward and expected policy entropy $\mathcal{H}(\pi(\cdot \mid \mathbf{s}_t))$ at each time step:
\begin{equation}
J(\pi) = \sum_{t=0}^{T} \mathbb{E}_{(s_t, a_t) \sim \rho_{\pi}} \left[\gamma^t( r(s_t, a_t) + \alpha \mathcal{H}(\pi(\cdot \mid \mathbf{s}_t))) \right]
\end{equation}
where $T$ is the terminal time step, $\gamma \in (0, 1]$ is the discount factor, and $\alpha$ is the temperature parameter, which determines the relative importance of the entropy term against the reward, and thus controls the stochasticity of the optimal policy \cite{haarnoja2018soft}. 
With this objective, the maximum entropy DRL methods have shown great potential in DRL efficient online exploration with sparse reward settings \cite{hiraoka2021dropout, ball2023efficient}, which fits the goal of this paper.

To apply maximum entropy RL in continuous spaces, one of the widely used methods, Soft Actor-Critic (SAC) \cite{haarnoja2018soft}, is applied.

\subsection{Imitation Bootstrapped Reinforcement Learning}
IBRL is a sample-efficient DRL framework that combines a standalone IL policy with an off-policy DRL policy \cite{IBRL}.
Firstly, IBRL requires an IL policy $\mu_{\psi}$ trained using expert demonstrations $\mathcal{D}$. The goal of $\mu_{\psi}$ is to mimic an expert behavior, and can be trained by minimizing a Behavior Cloning (BC) loss $\mathcal{L}_{BC}$:
\begin{align}
    \mathcal{L}_{B C}\left(\psi_{\mu}\right)=\mathbb{E}_{(s'
    , a') \sim \mathcal{D}}\left\|\mu_{\psi}(s')-a'\right\|_{2}^{2}
\end{align}
Then IBRL leverages the trained $\mu_{\psi}$ to help the DRL policy $\pi_{\theta}$ with online exploration and its target value estimation, referred to as the actor proposal phase and the bootstrap proposal phase respectively. In the actor proposal phase, the IBRL selects between an IL action, $a^{IL}\sim\mu_{\psi}(s_t)$, and an RL action, $a^{RL}\sim\pi_{\theta}(s_t)$. The one with a higher Q-value computed by the target critic networks, $ Q_{\phi^{\prime}}$, gets picked for the online interaction. Further to prevent local optimum Q-value update, the soft IBRL selects actions according to a Boltzmann distribution over Q-values instead of taking the $\operatorname{argmax}$:
\begin{align}
    a^{*}=\underset{a \in\left\{a^{\mathrm{IL}}, a^{\mathrm{RL}}\right\}}{\operatorname{argsoftmax}} Q_{\phi^{\prime}}(s, a) .
    \label{eqa:soft_a}
\end{align}

Similarly, in the Bootstrap Proposal phase, the future rollout will be carried out by selecting the action by $\operatorname{argmax}$ or $\operatorname{argsoftmax}$ between $Q_{\phi'}(s_{t+1}, a_{t+1}^{\text{IL}}) \quad \text{and} \quad Q_{\phi'}(s_{t+1}, a_{t+1}^{\text{RL}})$. The critic networks $Q_{\phi}(s_t,a_t)$ are updated as \eqref{eqa:IBRL_crtic_update}.
The RL policy network, $a^{RL}\sim\pi_{\theta}$, is updated the same as the selected off-policy DRL.

\section{Methods}

To reduce the exploration time wasted in correcting unreliable overestimated Q-values, and in turn improves exploration efficiency, it is crucial for the policy to favor distributions whose Q-values are more stable. This motivates selecting batches with reliable Q-value evaluations when updating both the critic and the policy networks.
Prior studies have shown that the Q-value estimates of $Q_{\phi'}\left(s_{t+1}, a(s_{t+1})\right)$ are only reliable when $(s_{t+1}, a(s_{t+1}))$ is from the same distributions as the dataset used to train $\hat{Q}(s_t, {a}_t)$ \cite{kumar2019stabilizing, nakamoto2023cal}. In our critic networks update process, instead of selecting between $Q_{\phi'}\left(s_{t+1}, \mu_{\psi}(s_{t+1})\right)$ and $Q_{\phi'}\left(s_{t+1}, \pi_{\theta}(s_{t+1})\right)$, where both $\left(s_{t+1}, \mu_{\psi}(s_{t+1})\right)$ and $\left(s_{t+1}, \pi_{\theta}(s_{t+1})\right)$ could be OOD state-action pairs. We propose to select between $Q_{\phi'}\left(s_{t+1}, \pi_{\theta}(s_{t+1})\right)$ and $Q_{\phi'}\left(s'_{t+1}, \mu_{\psi}(s'_{t+1})\right)$, where $s'_{t+1}$ are only sampled from $\mathcal{D}$.
This modification ensures $(s'_{t+1}, \mu_{\psi}(s'_{t+1}))$ is always from the same distribution as the $\mathcal{D}$, providing a reliable and calibrated Q-value estimates of the reference policy, whose values are on the similar scale as the true return value of $\mathcal{D}$ \cite{nakamoto2023cal}. 
With $\mathcal{D}$ fixed, we compare the mean estimated return of $\left(s_{t+1}, \pi_{\theta}(s_{t+1})\right)$ sampled from $\mathcal{B}$ against the bootstrapping-error–free ground-truth mean return of $\mathcal{D}$, thereby reducing the accumulated bootstrapping error in the action selection process.
Thus, \eqref{eqa.target} when updating the critic network becomes:
\begin{align}
    Q_{\phi}\left(s_{t}, a_{t}\right) \leftarrow r(s_{t}, a_{t})+\gamma Q_{\phi'}\left(s_{t+1}, \mathbf{a^{*}}(s_{t+1})\right)
    \label{eqa.target_est}
\end{align}
Compared with IBRL, the key modification is a simple action selection module, denoted as $\mathbf{a^*}(s)$:
\begin{align}
\mathbf{a^*}(s) = \begin{cases}
\mu_{\psi}(s), &\overline{Q}_{\phi^{\prime}}(s', \mu_{\psi}(s')) > \overline{Q}_{\phi^{\prime}}(s, \pi_{\theta}(s)), \\
\pi_{\theta}(s), & \text{otherwise}.
\end{cases}
\label{eqa.action_seclect}
\end{align}
where $\overline{Q}$ denote the mean of estimated Q-values, $s$ are the states from $\mathcal{B}$, and $s'$ are the states only sampled in the $\mathcal{D}$.

In the bootstrap proposal phase, the future rollouts $s_{t+1}$ are sampled randomly from $\mathcal{B}$. One can select $s'_{t+1}$ by finding the states closest to $s_{t+1}$ within $\mathcal{D}$, to enable more precise comparisons between nearby state–action pairs. However, for implementation simplicity, the current $s'_{t+1}$ is uniformly sampled from $\mathcal{D}$. By simple random sampling, the expected sample mean Q-value, $\overline{Q}_{\phi^{\prime}}(s', \mu_{\psi}(s'))$, from each batch converges to the population mean Q-value of the expert buffer \cite{rice2007mathematical}. Therefore, even though the comparison is made across different states, it remains valid because we are comparing the mean Q-values of the distributions induced by the IL policy and the RL policy.

Similarly, to align the policy strategy in the online interaction phase with the policy selected to propose future rollouts, the same action selection module \eqref{eqa.action_seclect} is used. With fewer OOD actions getting selected, the state distribution shift is also mitigated.
However, if $\mu_{\psi}(s)$ fails to provide good or recovery actions towards any state distribution shift, the considered action selection module might fail as $Q_{\psi^{\prime}}(s', \mu_{\psi}(s'))$ is not updated with fixed $\mathcal{D}$, and the same bad behavior from the reference policy might keep getting selected. Therefore, to leverage this action selection module for enhanced exploration efficiency, the initial online exploration states should lie within or near those in $\mathcal{D}$, and the reference policy should remain robust under small shifts in the state distribution.
 
Furthermore, to prevent the RL policy from overfitting the demonstration dataset and converging on a sub-optimal policy, we propose to replace TD3 with SAC. In SAC, the critic parameters $\phi$ are updated by minimizing the soft Bellman residual:
    \begin{equation}
    J_Q(\phi) = \mathbb{E}_{(s_t, a_t) \sim \mathcal{D}} \left[ \frac{1}{2} \left( Q_{\phi}(s_t, {a}_t) - \hat{Q}_{\phi}(s_t, {a}_t) \right)^2 \right],
    \label{crtic_update}
    \end{equation}
where ${Q}_{\phi}\left(s_{t}, a_{t}\right)$ is estimated using \eqref{eqa.target_est_sac}:

\begin{equation}
\begin{split}
Q_{\phi}\left(s_{t}, a_{t}\right)
& \leftarrow r(s_{t}, a_{t})+\gamma \bigr(Q_{\phi'}\left(s_{t+1}, \mathbf{a^{*}}(s_{t+1})\right) - \\
& \alpha\log \pi_{\theta}\bigl(f_{\theta}(\epsilon_{t+1}; s_{t+1})\mid s_{t+1})\bigr)
\label{eqa.target_est_sac}
\end{split}
\end{equation}

The stochastic actor parameters $\theta$ are updated by minimizing the expected KL-divergence: 
\begin{equation}
\begin{split}
J_{\pi}(\theta)
&= \mathbb{E}_{s_t \sim \mathcal{D},\,\epsilon_t \sim \mathcal{N}}
   \Bigl[\alpha\log \pi_{\theta}\bigl(f_{\theta}(\epsilon_t; s_t)\mid s_t\bigr) \\
&\qquad\quad {}- Q_{\phi}\bigl(s_t,\,f_{\theta}(\epsilon_t; s_t)\bigr)\Bigr]\,. 
\label{actor_update}
\end{split}
\end{equation}
where the stochastic action is $f_{\theta}(\epsilon_t; s_t)$, and $\epsilon_t$ is an input noise distribution, sampled from some fixed distribution \cite{haarnoja2018soft}. We propose that the distribution can be the demonstration $\mathcal{D}$, but in this study, we only consider a simple Gaussian distribution $\mathcal{N}$. $\log \pi_{\theta}(f_{\theta}(\epsilon_t; s_t) \mid s_t) $ is the log‑probability of the stochastic action $f_{\theta}(\epsilon_t; s_t)$ under the current policy $\pi_{\theta}$. 

Lastly, to leverage the robustness of the proposed framework towards the quality of the demonstration. 
% Before the RL training starts, the first modification is to leverage the quality of the demonstration. 
We propose to choose offline DRL as the reference policy (IL policy in the IBRL framework) when the quality of the demonstration is unknown or imperfect. With strong sequential decision-making ability, offline DRL can be more robust to the demonstration quality than IL methods \cite{kumar2020conservative, fujimoto2021minimalist}.

Combining all the modifications, the \IBRLn{} is introduced in Algorithm~\ref{alg:ibrln}, our new modifications are marked in red. 
\begin{algorithm}
\caption{\IBRLn}
\label{alg:ibrln}
\begin{algorithmic}[1]
\STATE \textbf{Input:} Critic networks $Q_{\phi_i}\left(s, a\right)$ and target critic networks $Q_{\phi_i'}$ with random initial parameter values; policy network $\pi_{\theta}$ and target policy network $\pi_{\theta'}$;
\STATE Initialize replay buffer $\mathcal{B}$ \textcolor{red}{and expert buffer $\mathcal{D}$};
\STATE Train an reference policy $\mu_{\psi}$ with expert buffer $D$ by IL or \textcolor{red}{offline RL}.
\FOR{each episode $M$}
    \STATE Reset environment to initial state $s_0$.
    \FOR{each time step $t$}
        \STATE Observe $s_t$ from the environment, compute IL action $a^{IL} \sim \mu_{\psi}(s_t)$ and \textcolor{red}{RL stochastic action $a^{RL} \sim \pi_{\theta}(f_{\theta}(\epsilon_t; s_t) \mid s_t)$}
        \STATE Compute Q-value from the target critic networks $Q_{\phi_i'}$.
        \STATE \textcolor{red}{Execute $\mathbf{a^*}$ based on \eqref{eqa.action_seclect}}.
        \STATE  Store transition $(s_t, a_t, r_t, s_{t+1})$ in replay buffer $\mathcal{B}$.
        \STATE Randomly sample a minibatch of $N$ transitions respectively from the replay buffer $\mathcal{B}$ and \textcolor{red}{$\mathcal{D}$}.
        \STATE Update critic networks parameters by \textcolor{red}{\eqref{crtic_update}}.
        \STATE Update actor networks parameters by \textcolor{red}{\eqref{actor_update}}.
        \STATE Update target networks.
    \ENDFOR
\ENDFOR
\end{algorithmic}
\end{algorithm}

\section{Experiment Design and Evaluation}
In this section, experiments are designed and conducted in the simulation to evaluate the proposed method. The experimental design and evaluation aim to answer the following core questions:

\subsection{How generalizable is \IBRLn{} across environments with varying reward densities and state-action space complexities?}
To answer the question, the tasks selected in the problem statement are studied under both dense reward and sparse reward settings. For the bucket loading task, the state and action dimensions are 4 and 3, respectively. The details, such as reward design, domain randomization, and prior demonstration collection, are covered in Section.~\ref{application}.
For the open drawer task, the state and action dimensions are 23 and 9, respectively. The details of the open drawer task are covered in \cite{cab_open}. The original reward design for the open drawer task is dense and contains: distance reward, open drawer reward, and some bonus reward for opening the drawer properly. To study the same task with a sparse reward setting, we simply set the distance reward gain to 0. To collect simulated demonstrations for the open drawer task, a TD3 policy was trained with dense, human-designed rewards. A total of 30 prior trajectories are recorded by evaluating the trained TD3 with random noise added to the policy output.

Both tasks are trained with Isaac Gym \cite{makoviychuk2021isaac}. All experiments with the open drawer task were run with 10 parallel environments, using two different random seeds (10 and 11) to ensure robustness and reproducibility. All experiments with the bucket loading task were run with a single environment, using two different random seeds (10 and 11). The detailed configurations for training each task are shown in Section.~\ref{Sec.config}.

Question A is answered through the following evaluation results: Figure~\ref{fig:exp3} demonstrates the performance of \IBRLn{} to learn the open drawer task with both sparse and dense reward, by achieving the highest reward in both reward settings, the results validate the robustness of \IBRLn{} towards varying reward densities. Figure~\ref{fig:exp3} and Figure~\ref{fig:aem_exp3} present the performance of \IBRLn{} with different state-action spaces complexity. By outperforming IBRL on the open drawer task and achieving comparable reward in the bucket loading task, the results validate the ability of \IBRLn{} to generalize across varying levels of state–action space complexity.

\subsection{How effective is the proposed action selection module in addressing the bootstrapping error and improving exploration efficiency during learning compared to IBRL?}
\begin{figure*}[ht]
  \centering
  \begin{subfigure}[b]{0.23\textwidth}
    \centering
    \includegraphics[width=\linewidth]{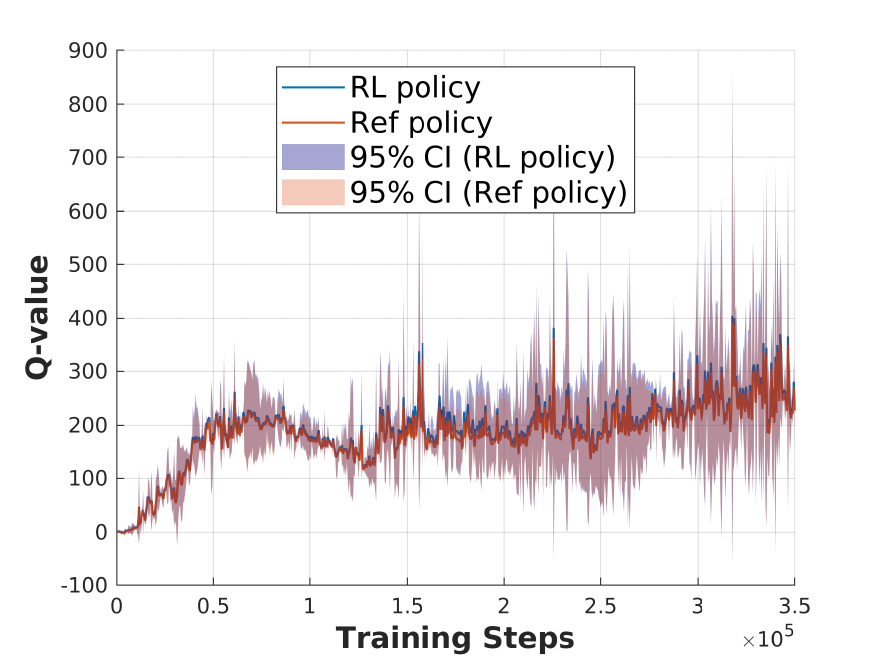}
    \caption{Q-value estimation with original IBRL.}
    \label{fig:exp2_1}
  \end{subfigure}
  \hfill
  \begin{subfigure}[b]{0.23\textwidth}
    \centering
    \includegraphics[width=\linewidth]{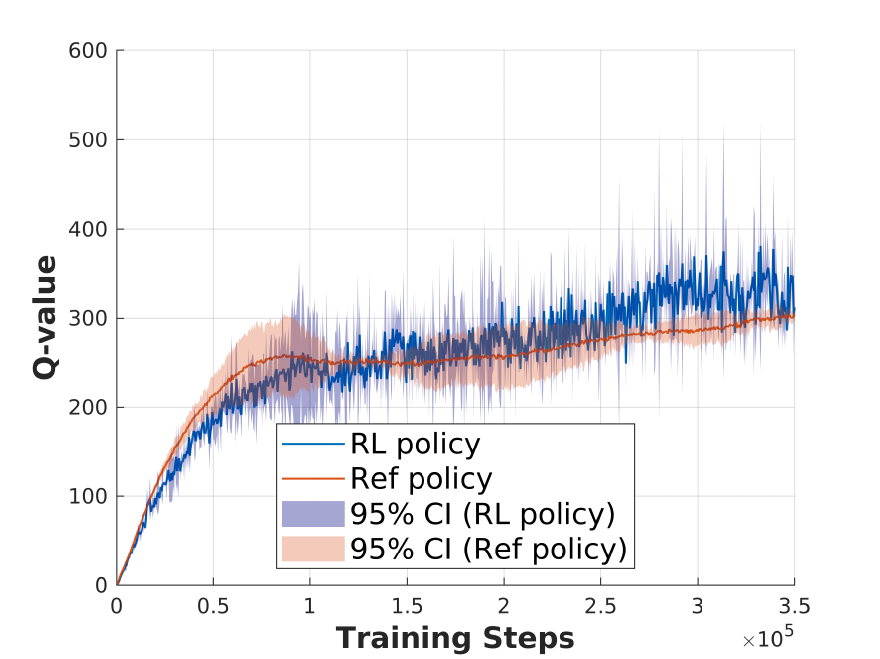}
    \caption{Q-value estimation with our proposed action selection module.}
    \label{fig:exp2_2}
  \end{subfigure}
  \hfill
  \begin{subfigure}[b]{0.23\textwidth}
    \centering
    \includegraphics[width=\linewidth]{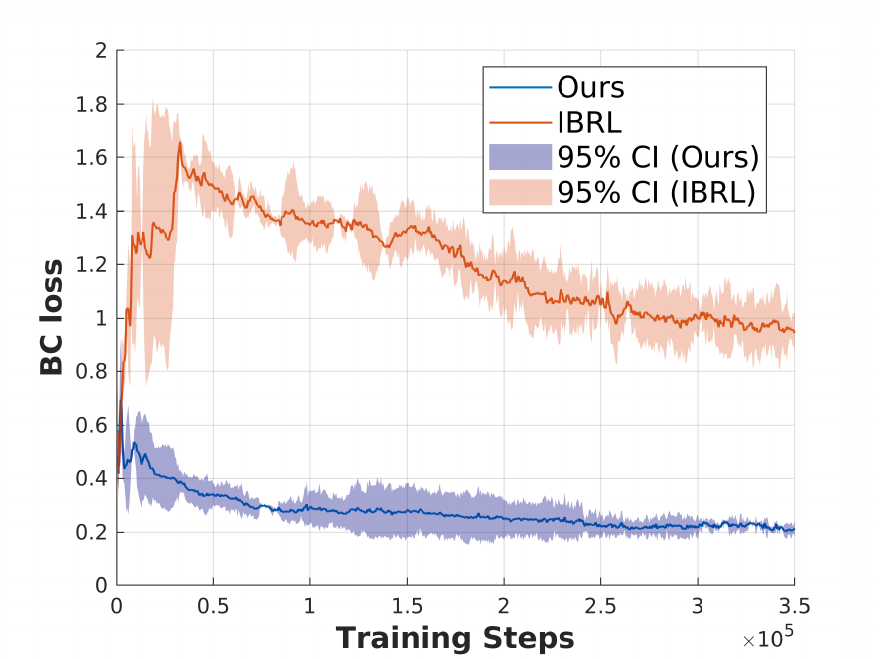}
    \caption{BC loss.}
    \label{fig:exp2_3}
  \end{subfigure}
  \hfill
  \begin{subfigure}[b]{0.23\textwidth}
    \centering
    \includegraphics[width=\linewidth]{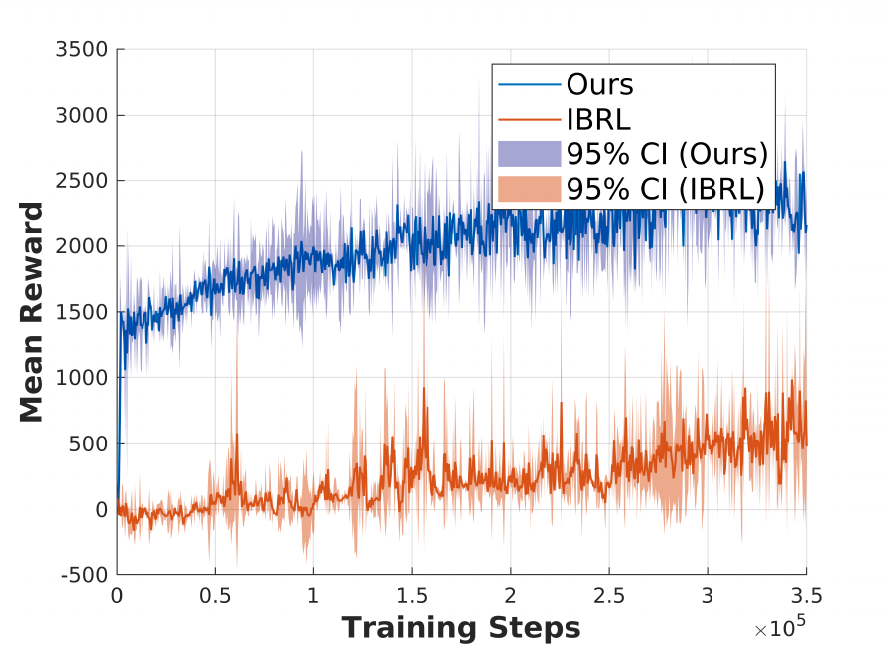}
    \caption{Mean reward.}
    \label{fig:exp2_4}
  \end{subfigure}
  \caption{Exp2:   the effectiveness of the proposed new action selection method with the Open Drawer task.}
  \label{fig:exp2}
\end{figure*}

\begin{figure*}[ht]
  \centering
  \begin{subfigure}[b]{0.23\textwidth}
    \centering
    \includegraphics[width=\linewidth]{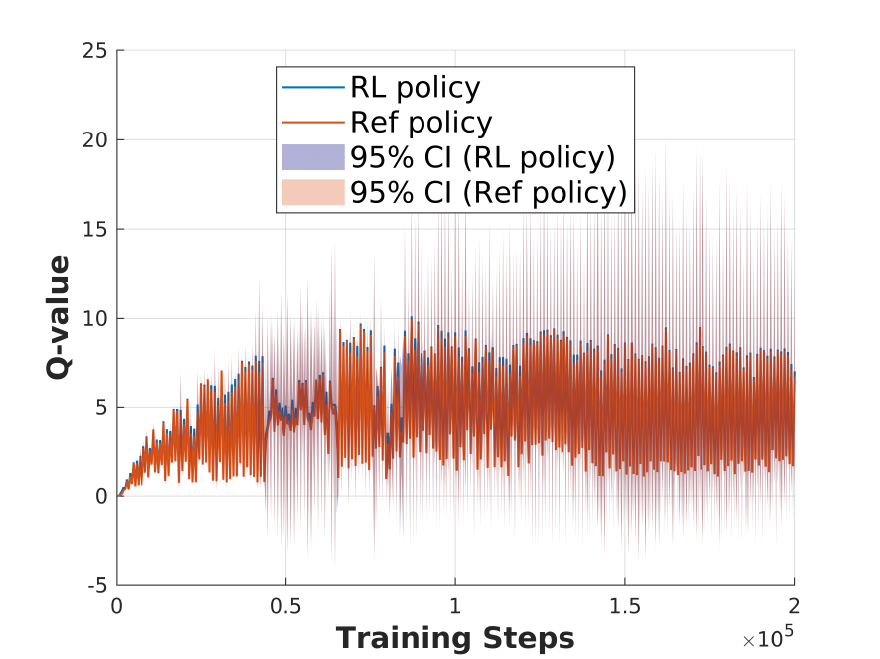}
    \caption{Q-value estimation with original IBRL.}
    \label{fig:exp2_5}
  \end{subfigure}
  \hfill
  \begin{subfigure}[b]{0.23\textwidth}
    \centering
    \includegraphics[width=\linewidth]{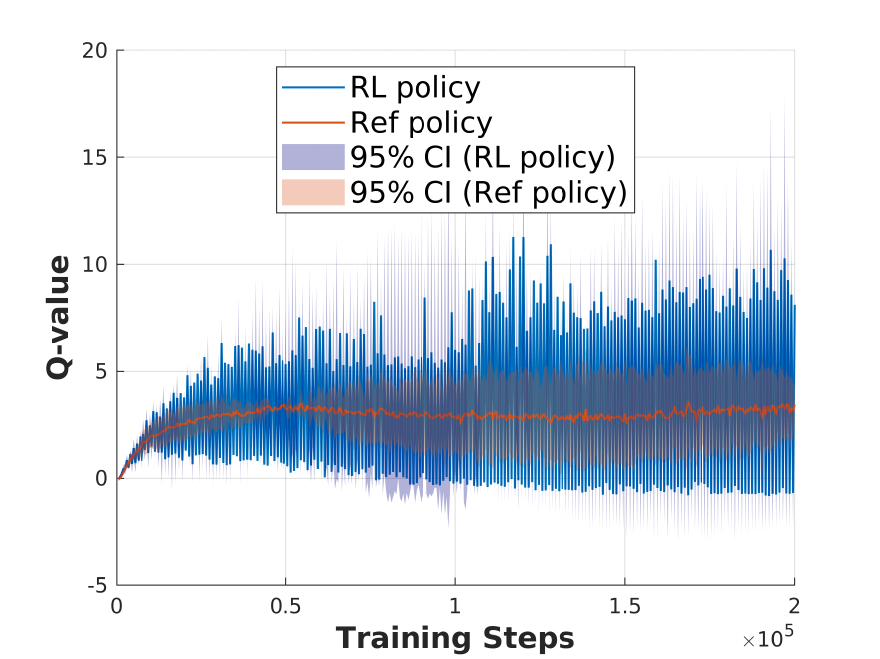}
    \caption{Q-value estimation with proposed new IBRL.}
    \label{fig:exp2_6}
  \end{subfigure}
  \hfill
  \begin{subfigure}[b]{0.23\textwidth}
    \centering
    \includegraphics[width=\linewidth]{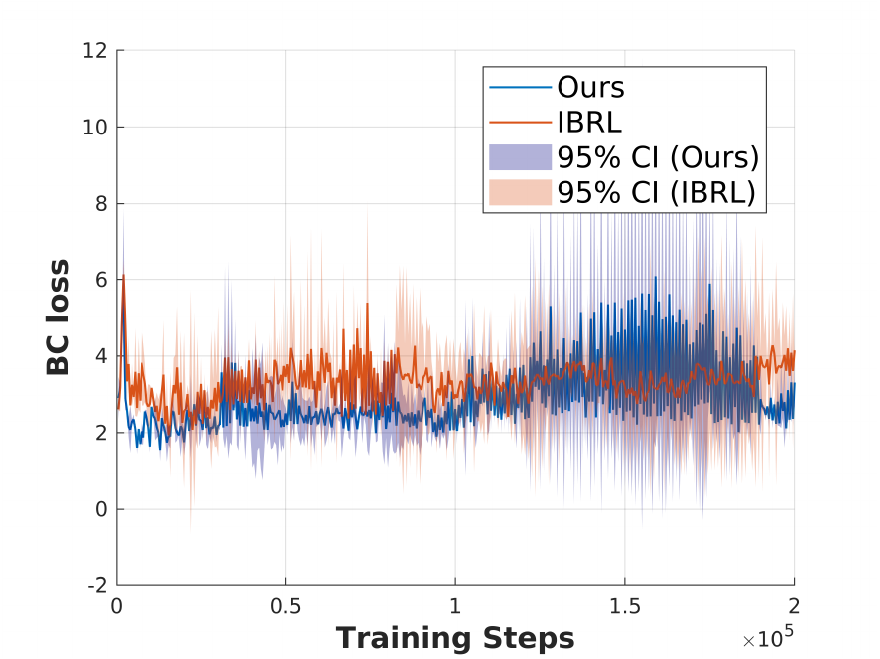}
    \caption{BC loss.}
    \label{fig:exp2_7}
  \end{subfigure}
  \hfill
  \begin{subfigure}[b]{0.23\textwidth}
    \centering
    \includegraphics[width=\linewidth]{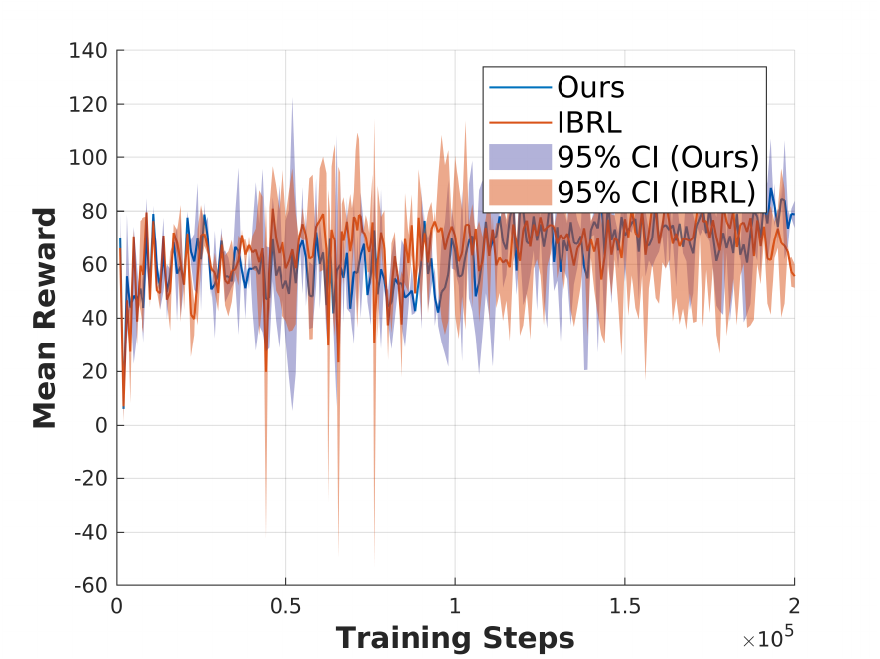}
    \caption{Reward.}
    \label{fig:exp2_8}
  \end{subfigure}
  \caption{Exp3: Validate the effectiveness of the proposed new action selection method with the Bucket Loading task.}
  \label{fig:exp2_aem}
\end{figure*}

To examine the effectiveness of the action selection module in addressing bootstrapping error and improving exploration efficiency, we conducted experiments in which only the action selection module of the original IBRL framework was replaced. The reference policy used is the IL policy, while the RL policy remains TD3 in both setups. 
Four criteria are recorded during training: 1) The Q-value of the Ref policy during action selection in the online interaction phase. 2) The Q-value of the RL policy during action selection in the online interaction phase. 3) BC loss: $\mathcal{L}_{B C}\left(\pi_{\theta}\right)=\mathbb{E}_{(s
, a) \sim \mathcal{B}}\left\|\pi_{\theta}(s)-a\right\|_{2}^{2}$, for measuring the difference between sampled actions in the replay buffer and the actions output by the RL policy. 4) Reward convergence over training steps. Figure~\ref{fig:exp2} and Figure~\ref{fig:exp2_aem} present a comparison of the considered criteria between the baseline IBRL and our proposed method across two selected tasks with the sparse reward setting.

The results for the open drawer task are shown in Figure~\ref{fig:exp2}. In Figure~\ref{fig:exp2_1}, we compared the Q-value of the Ref policy and RL policy during action selection in the online interaction phase in the IBRL. The Q-values of the Ref policy appear closely estimated to the RL policy, and both of the Q-values have high variances during the training. Combine the results of the BC loss between sampled actions and the agent’s output actions in Figure~\ref{fig:exp2_3}, indicating a mismatch between the updated policy and the behavior policy, suggesting the OOD actions are getting selected due to the bootstrapping error discussed in Section.~\ref{problem_state}. As a result, the Ref policy failed to get selected to provide reliable guidance, as reflected in the degraded performance in Figure~\ref{fig:exp2_4}.
While Figure~\ref{fig:exp2_2} presents a stable Q-value estimation of the Ref policy, and a clear higher mean value compared with the RL policy in the early training steps, which aligns with the core idea of the IBRL framework. The corresponding BC loss in Figure~\ref{fig:exp2_3} is significantly reduced by approximately $80\%$ compared to the BC loss of IBRL, indicating the bootstrapping error is effectively mitigated with our action selection method. Consequently, the Ref policy succeeded in efficient guidance throughout the RL training, as demonstrated by the improved reward convergence in Figure~\ref{fig:exp2_4}. The proposed action selection module achieved a mean reward approximately four times higher than IBRL during the interaction steps.

The results for the bucket loading task are shown in Figure~\ref{fig:exp2_aem}. Notably, the experiments of the bucket loading task were run with a single environment since it is computationally expensive to simulate thousands of particles in parallel environments. Thus, the results of the bucket loading tasks have higher variance compared to the open drawer task, where 10 environments are running in parallel. The results suggest the action selection module has less effect on the low-dimensional state-action task, and the original IBRL can already score a near-optimal reward. This can also be attributed to the performance of the Ref policy. If the RL policy can easily acquire a higher Q-value than the Ref policy, the effect of our action selection module will be limited.
Nevertheless, the stable Q-value estimation of the Ref policy in Figure~\ref{fig:exp2_6} still validates the effectiveness of our action selection module in maintaining reliable Q-value estimations.

\subsection{How effective is SAC in improving exploration efficiency during learning compared to the initial IBRL?}
To examine the effectiveness of the SAC in improving exploration efficiency, we conducted experiments: 1) The original IBRL, denoted as $\text{IBRL}_{TD3}$. 2) The IBRL with our action selection module, denoted as $\text{Ours}_{TD3}$. 3) The IBRL with SAC to be the RL policy, denoted as $\text{IBRL}_{SAC}$. 4) Our \IBRLn{} framework, denoted as $\text{Ours}_{SAC}$. The Ref policy remains IL policy in all setups.

\begin{figure}[htb]
  \centering
  \begin{subfigure}[b]{0.5\columnwidth}
    \centering
    \includegraphics[width=\linewidth]{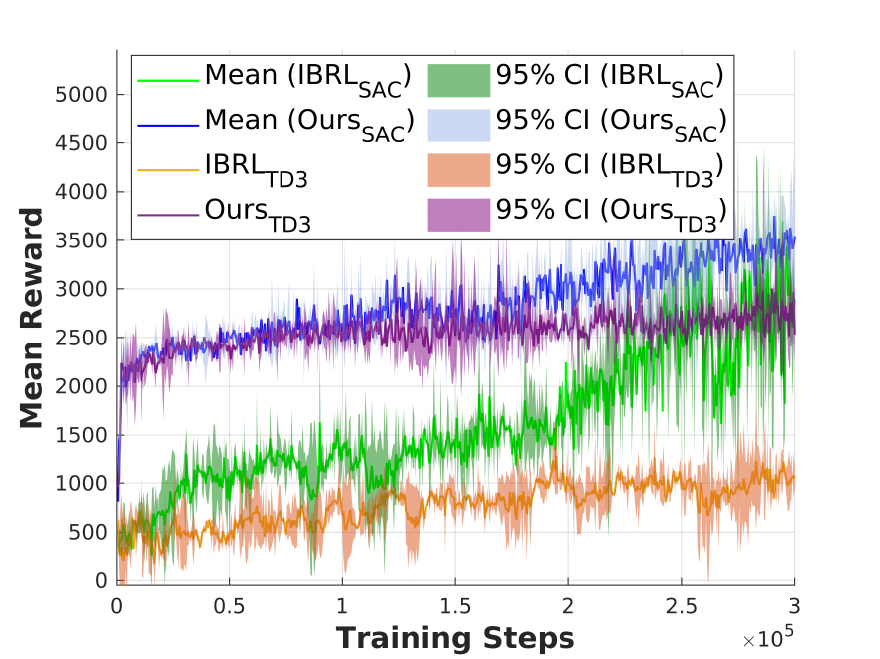}
    \caption{Dense reward setting.}
    \label{fig:exp3_1}
  \end{subfigure}
  \hfill
  \begin{subfigure}[b]{0.48\columnwidth}
    \centering
    \includegraphics[width=\linewidth]{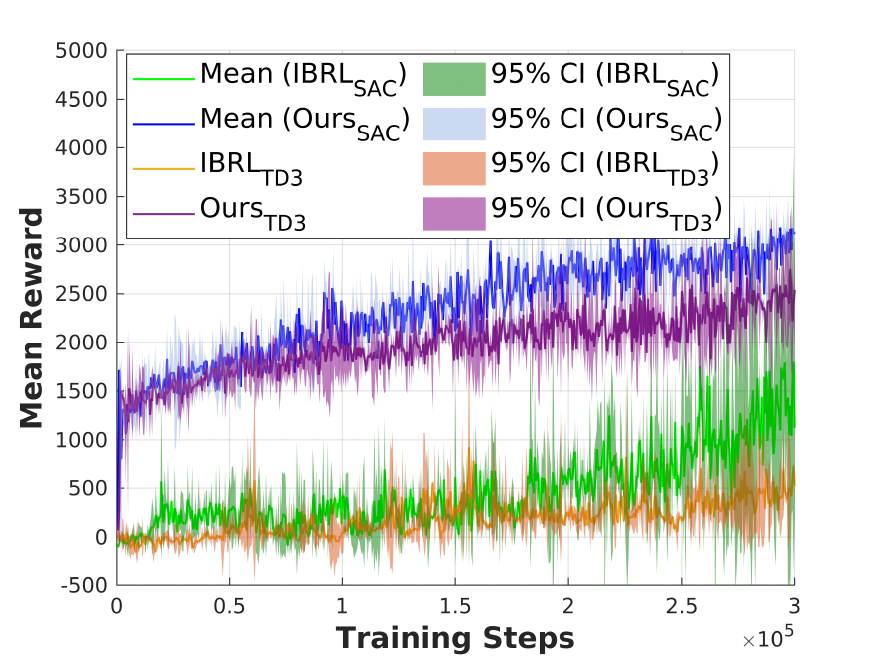}
    \caption{Sparse reward setting.}
    \label{fig:exp3_2}
  \end{subfigure}
  \caption{Exp4: Validate the effectiveness of SAC with the open drawer task. }
  \label{fig:exp3}
\end{figure}

\begin{figure}[htb]
    \centering
    \includegraphics[width=0.8\linewidth]{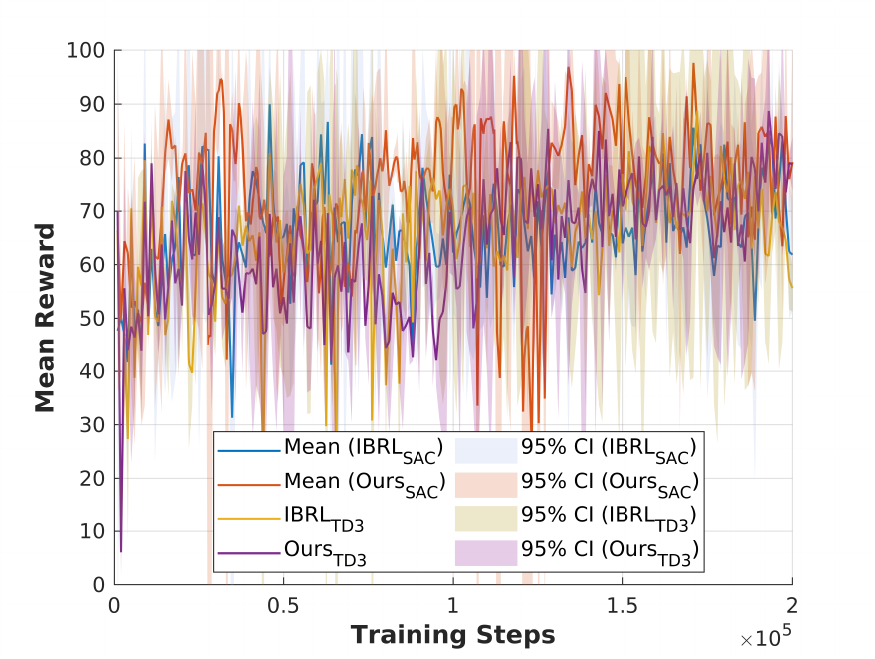}
    \label{fig:exp3_3}
  \caption{Exp5: Validate the effectiveness of SAC with the bucket loading task. }
  \label{fig:aem_exp3}
\end{figure}

The reward convergence over training steps is recorded as the main evaluation criteria. Figure~\ref{fig:exp3} and Figure~\ref{fig:aem_exp3} present a comparison of the considered experiments across two selected tasks.
The results for the open drawer task with varied reward settings are shown in Figure~\ref{fig:exp3}.
The reward convergence suggests that, with the same training steps, the experiments with $SAC$ are able to explore higher rewards compared to experiments with $TD3$, which converged on a sub-optimal reward.
The results for the bucket loading task with sparse reward settings are shown in Figure~\ref{fig:aem_exp3}. The results suggest our method and the IBRL achieve similar performance in low-dimensional state-action spaces.

The final evaluation results of each algorithm across two tasks are shown in Table.~\ref{tab:eval_reward}. Table.~\ref{tab:eval_reward} shows that \IBRLn{} achieves the best evaluation performance in both tasks. In the open drawer task with sparse reward setting, \IBRLn{} improves the averaged reward by around $347\%$, showing a dramatic improvement.
\begin{table}[ht!]
\centering
\setlength{\tabcolsep}{4pt} % Column spacing
\renewcommand{\arraystretch}{1.1} % Row spacing
\scriptsize % Reduce font size to fit
\begin{tabular}{lcccc}
\hline
Task & $\text{IBRL}_{TD3}$ & $\text{Ours}_{TD3}$ &  $\text{IBRL}_{SAC}$ &  $\text{Ours}_{SAC}$ \\
\hline
open drawer (dense) & 1055 & 2735 & 2747 & 3455\\
open drawer (sparse) & 682.6 & 2475 & 2150 & 3053 \\
bucket loading (sparse) & 71.7 & 76.5 & 69.9 & 81.8 \\
\hline
\end{tabular}
\caption{Averaged rewards of evaluating each RL policy at the last time step over 5 episodes.}
\label{tab:eval_reward}
\end{table}

\subsection{What is the impact of demonstration quality on the performance of our method?}
\begin{figure}[htb]
  \centering
  \begin{subfigure}[b]{0.45\columnwidth}
    \centering
    \includegraphics[width=\linewidth]{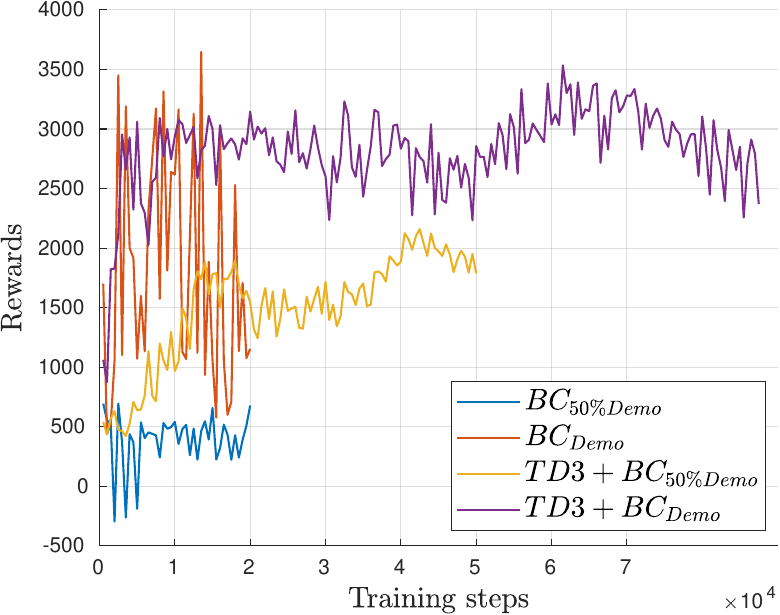}
    \caption{Comparison between IL policy and offline RL policy.}
    \label{fig:exp4_1}
  \end{subfigure}
  \hfill
  \begin{subfigure}[b]{0.45\columnwidth}
    \centering
    \includegraphics[width=\linewidth]{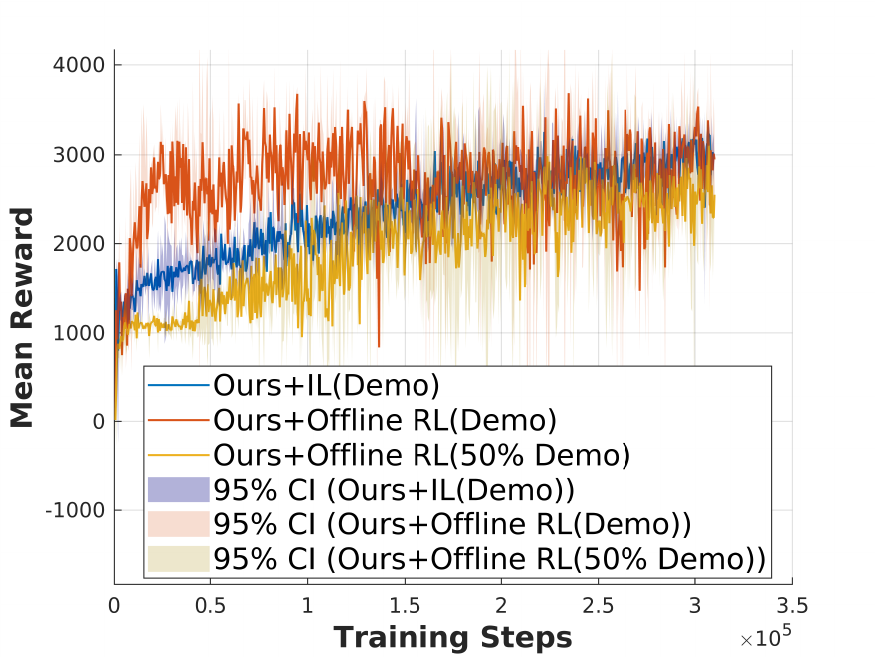}
    \caption{Reward convergence with the varying demonstration qualities.}
    \label{fig:exp4_2}
  \end{subfigure}
  \caption{Exp6: Validate the robustness of our framework towards the quality of demonstration with the open drawer task.}
  \label{fig:exp4}
\end{figure}

To evaluate the robustness of the proposed method to demonstration quality, the following experiments were conducted: we fill the demonstration dataset with 1) 50 \% data from the random policy, denoted as $50\% demo$. 2) Suboptimal demo: Add noise to the expert policy outputs. 
For simplicity, a BC policy is selected as the IL policy. A minimalist approach to offline RL, known as TD3+BC \cite{fujimoto2021minimalist}, is selected as our Ref policy. For the sake of the complexity in designing such experiments, only the open drawer task with sparse reward, which is the most difficult to learn, is evaluated in the experiments.
The results are shown in Figure~\ref{fig:exp4}. Figure~\ref{fig:exp4_1} demonstrated that TD3+BC can learn a good policy even from $50\% demo$, while BC failed. Furthermore, TD3+BC also learns a better policy using the suboptimal demo. Figure~\ref{fig:exp4_2} validated the robustness of our method towards varying demonstration quality, by achieving the same level of rewards with both datasets. 

% \subsection{How does our method perform in comparison to other approaches that utilize prior demonstrations?}
% To examine the effectiveness of our method compared to other methods with prior demonstrations. We compared the interaction steps required for the reward convergence with the following baselines.
% \begin{figure}[htb]
%   \centering
%   \begin{subfigure}[b]{0.45\columnwidth}
%     \centering
%     \includegraphics[width=\linewidth]{images/IBRL-SAC1.pdf}
%     \caption{Dense reward setting.}
%     \label{fig:1}
%   \end{subfigure}
%   \hfill
%   \begin{subfigure}[b]{0.48\columnwidth}
%     \centering
%     \includegraphics[width=\linewidth]{images/IBRL-SAC2.pdf}
%     \caption{Sparse reward setting.}
%     \label{fig:3}
%   \end{subfigure}
%   \caption{Comparison of mean reward convergence with two tasks in sparse reward setting.}
%   \label{fig:examples}
% \end{figure}

To this end, we have demonstrated the effectiveness of the proposed method. The method is also applied to a real industrial application to showcase the implementation process and the sim2real performance.

\section{Real industrial applications}

\begin{figure*}[ht]
   \centering   \includegraphics[width=0.7\textwidth]{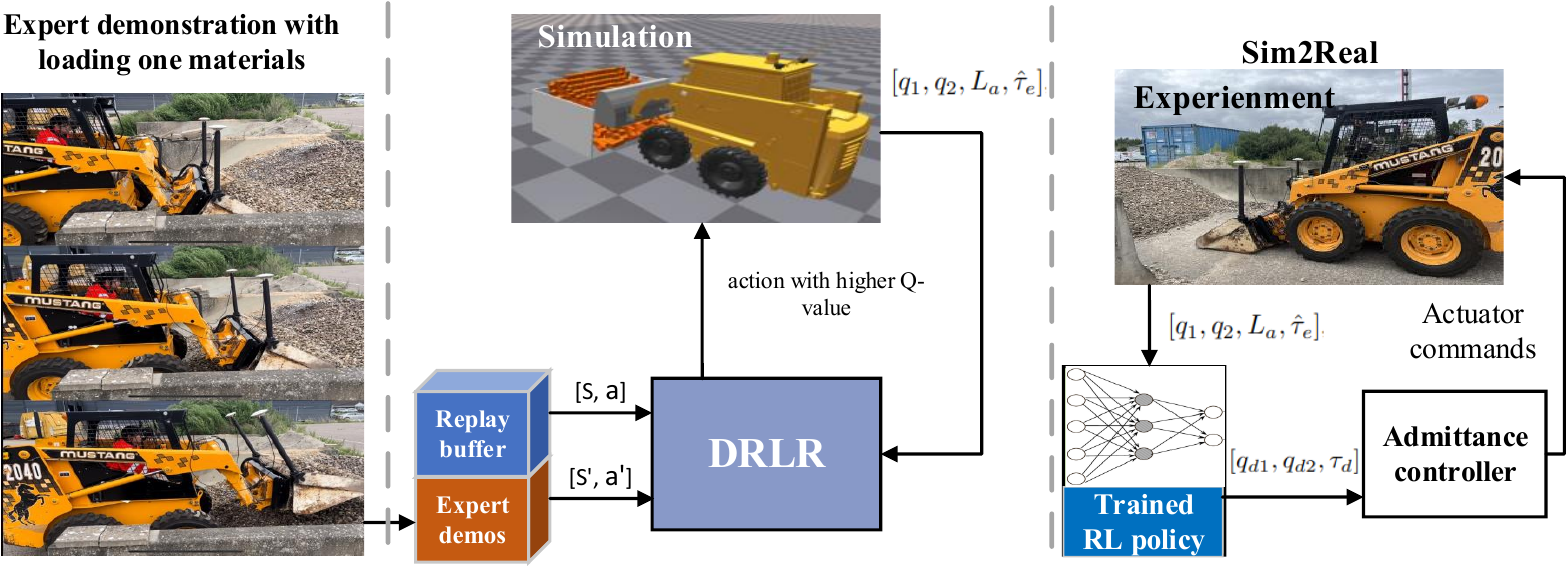}
   \caption{Illustration of the implementation of applying the proposed framework to the automatic wheel loader loading task.}
   \label{fig:framework}
\end{figure*}
This section presents an application of the proposed framework for the wheel loader loading task, where only a limited number of expert demonstrations are given to demonstrate the data efficiency. The detailed implementation is illustrated in Fig.~\ref{fig:framework}. 

\subsection{Bucket-media simulation}
Before learning with the proposed framework, it is important to create an environment similar to the real world to enable policy exploration, while applying domain randomization to deal with observation shifts. In the simulation, the wheel loader is configured with the same dynamic parameters obtained from a real machine. 
Because it is impractical to directly model the hydraulic actuation force or the bucket-media interaction force under different materials and geometries, this paper attempts to regularize the external torque rather than model it. We proposed to use admittance controllers to decrease the variances in the external torque by changing the position reference. 
The implementation of the admittance controller is given in the Appendices.

\begin{figure}[htp]
    \centering
    \includegraphics[width=8.0cm]{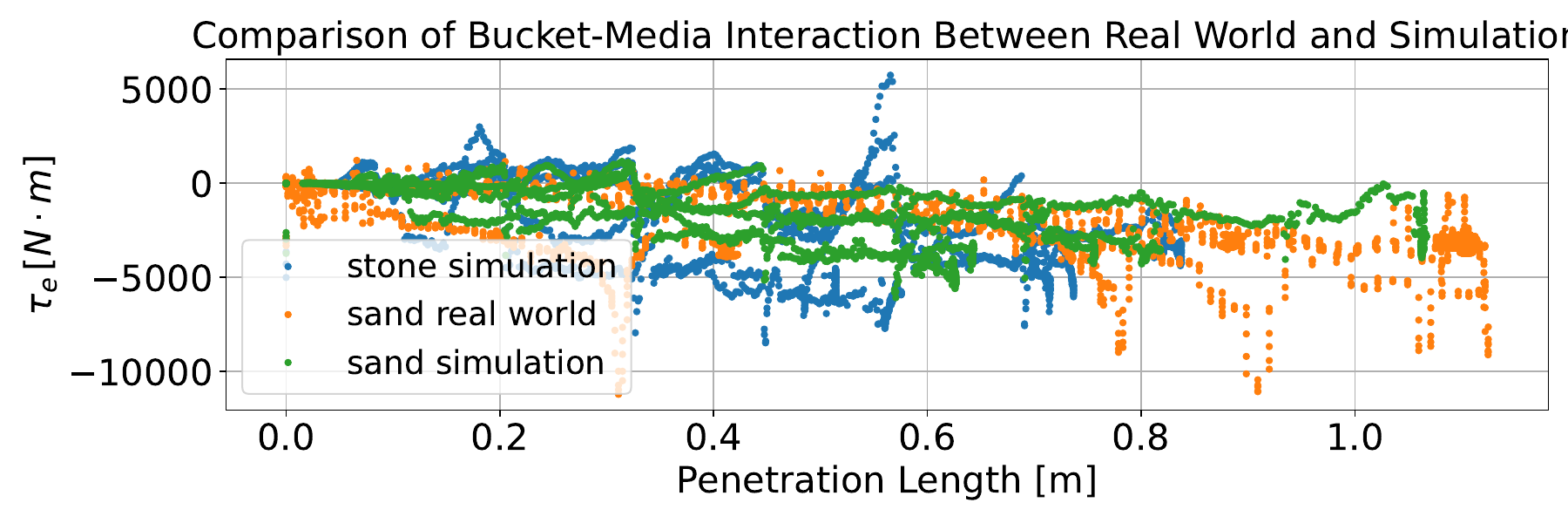 }
    \caption{Comparison of estimated external torque during penetrating between simulation and the real world. In the real world (orange), the external torque is measured while loading dry sand. In the simulation, the external torque (green and blue) is generated by loading sand and stone piles, using the same penetration motion in the real-world experiment. }
    \label{fig:simexp_comp}
\end{figure}

Table~\ref{tab:domran} shows the parameters we randomized to simulate bucket-media interactions with different pile geometries and pile materials. A comparison of the estimated external torque during penetrating the pile between simulation and the real world is presented in Fig.~\ref{fig:simexp_comp}. Different from real-world settings, the external torque is estimated from contact sensors in the simulation, due to the poor performance of the force sensor in IsaacGym.

\begin{table}[ht!]
  \centering
\begin{tabular}{ccccc}
  \hline
  Domain randomization & Range \\\hline
  density & $[1700\pm100 \quad 2600\pm100] kg/m^3$ \\
  pile geometry & [$25^{\circ}$, $45^{\circ}$,  $55^{\circ}$] \\
  particle friction & $[0.3\quad 0.4]$ \\
  white noises on observations & [-1e-4 1e-4] \\
  \hline
\end{tabular}
\caption{Domain randomization parameters and their sampling ranges.}
\label{tab:domran}
\end{table}

\subsection{\IBRLn{} implementation}
Both the Ref and RL policies have $4$ inputs: $[q_1, q_2, L_a, \hat{\tau}_e]$, representing: boom joint position, bucket joint position, advancing length, estimated external torque; and $3$ outputs: $[q_{d1}, q_{d2}, \tau_d]$, where $q_{d1}, q_{d2}$ are desired position references for boom and bucket joint, and $\tau_d$ is desired torque reference for admittance control that is only used during penetrating.

\begin{figure}[H]
    \centering
    \includegraphics[width=8.0cm]{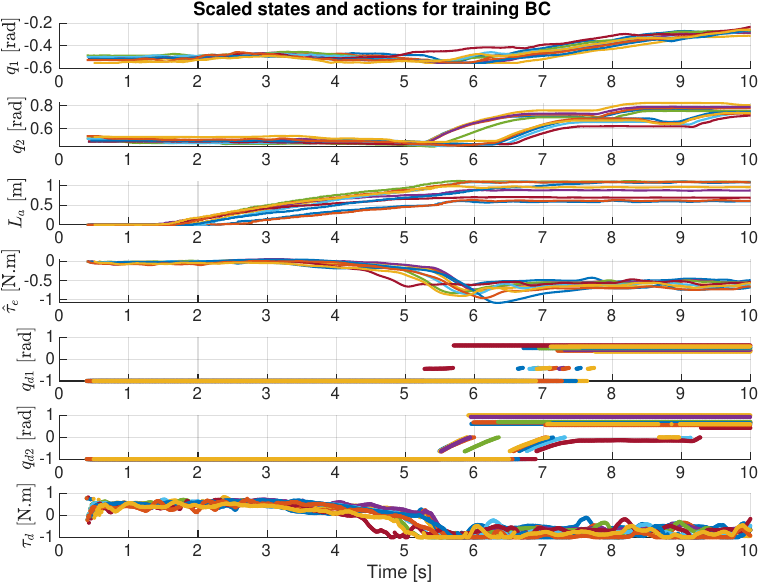}
    \caption{States-actions pairs for training the Ref policy. Each curve represents the data recorded in one bucket loading demonstration.}
    \label{fig:states-actions}
\end{figure}

To train the Ref policy, 10 expert demonstrations of loading dry sand piles with changing pile geometries are recorded.
In the demonstration, $[q_1, q_2, L_a]$ are directly used as inputs, $\hat{\tau}_e$ is scaled between $[-1, 1]$. Position references are acquired from forward dynamics of sent actuation signals from the demonstrations, they are firstly normalized and then used as $[q_{d1}, q_{d2}]$, scaled $\hat{\tau}_e$ is directly assigned as $\tau_d$. The state-action pairs that are used for training the reference policy are shown in Fig.~\ref{fig:states-actions}. For simplicity, BC is employed to train the reference policy.

% According to \cite{Eriksson2023}, compared to Stacked Multi-Layer Perceptron (MLP-S), Long short-term memory (LSTM), and Random forest (RF), Multi-Layer Perceptron (MLP) has the best performance in imitation learning for wheel loader loading tasks in terms of loading efficiency and robustness. However, in their work, more than 50 sets of bucket-filling demonstrations are used to train the MLP.
% In contrast, this study is limited to only 10 sets of expert demonstrations, aiming to extract the maximum information from the available data. The BNN, which has not yet been investigated for learning loading tasks, offers additional uncertainty estimation for the predicted actions. 
% Therefore, BNN is selected for training the BC policy. BNN used in this work consists of two hidden layers for more robust performance \cite{Yang2021}, with 64 and 32 neurons respectively. The Exponential Linear Unit (ELU) activation function is applied to each layer, and a dropout rate of 0.3 is used to mitigate overfitting.

\begin{figure}[htp]
    \centering    \includegraphics[width=7.0cm]{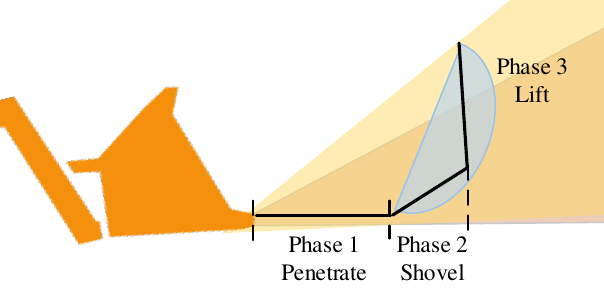}
    \caption{Three phases for the wheel loader loading process. }
    \label{fig:phase}
\end{figure}

The wheel loader loading process can be divided into 3 phases as shown in Fig.~\ref{fig:phase}: penetrate, shovel, and lift \cite{Sarata2004}.
To train the DRL, the bucket loading task is divided into two sub-tasks as shown in \eqref{eqa:phase}. 
\begin{equation}
    subtask =\left\{\begin{array}{ll}
    P_1 ,& q_{d2} > -0.5\\
    P_2 \& P_3 & else
    \end{array}\right.
    \label{eqa:phase}
\end{equation}
In phase 1: $P_1$, the boom and bucket penetrate the pile with an admittance controller tracking $q_{d1}, q_{d2},\tau_d$, and the loader moves forward with a constant velocity. In phase 2 and 3: $P_2 \& P_3$, the controller switches to an inverse dynamics controller with only tracking the position references $q_{d1}, q_{d2}$, and the loader stops moving forward. The transition between $P_1$ and $P_2\& P_3$ is determined by when the loader stops moving forward. Based on observing the demonstrations, this transition is identified when the desired bucket reference position $q_{d2}$ surpasses approximately -0.5.
 
The goal for the bucket loading task is to achieve a full bucket-fill rate, and the boom-bucket joint reaching its designated end position, corresponding to the maximum allowable value within the position reference range. This leads to a natural sparse reward setting, where the reward only occurs at the end of the tasks. However, sparse reward requires a longer training time because it is more difficult for the RL agent to explore than dense reward settings. Although previous work \cite{Shen2024} demonstrated a successful performance with dense reward setting, designing such rewards is challenging and may lead to suboptimal actions.
Since our framework has shown robust performance in sparse reward settings, a simpler sparse reward setting is designed as follows: 
\begin{equation}
    r =\left\{\begin{array}{ll}
    R_{f} + R_{e} ,& T-50 \\
    - 10, & Fail\\
    0,& Else
    \end{array}\right.
\end{equation}
where $T$ represents the final step of an episode. A failure of loading ($Fail$) occurs if the bucket fill rate reward, $R_{f}$, and the end reward, $R_{e}$, do not achieve at least half of their maximum designed values by the end step $T$. The rewards $R_{f}$ and $R_{e}$ are defined as follows:
\begin{equation}
    R_{f} = \frac{V}{V_{max}} , \quad
    R_{e} = 1 - \frac{d}{d_{max}}. 
\end{equation}
where $V_{max}$ is the bucket capacity, $V$ is current bucket load volume, and $V = \hat{\tau}_e/\rho_{rad}gl_1$, where $\rho_{rad}$ is the particle density, $l_1$ is the length of boom. $d$ is the Euclidean distance between the current boom, bucket joint position to the end position, $d_{max}$ is the Euclidean distance between the initial boom, bucket joint position to the end position.

\subsection{Sim2real results}
\begin{figure}[htp]
    \centering    \includegraphics[width=7.0cm]{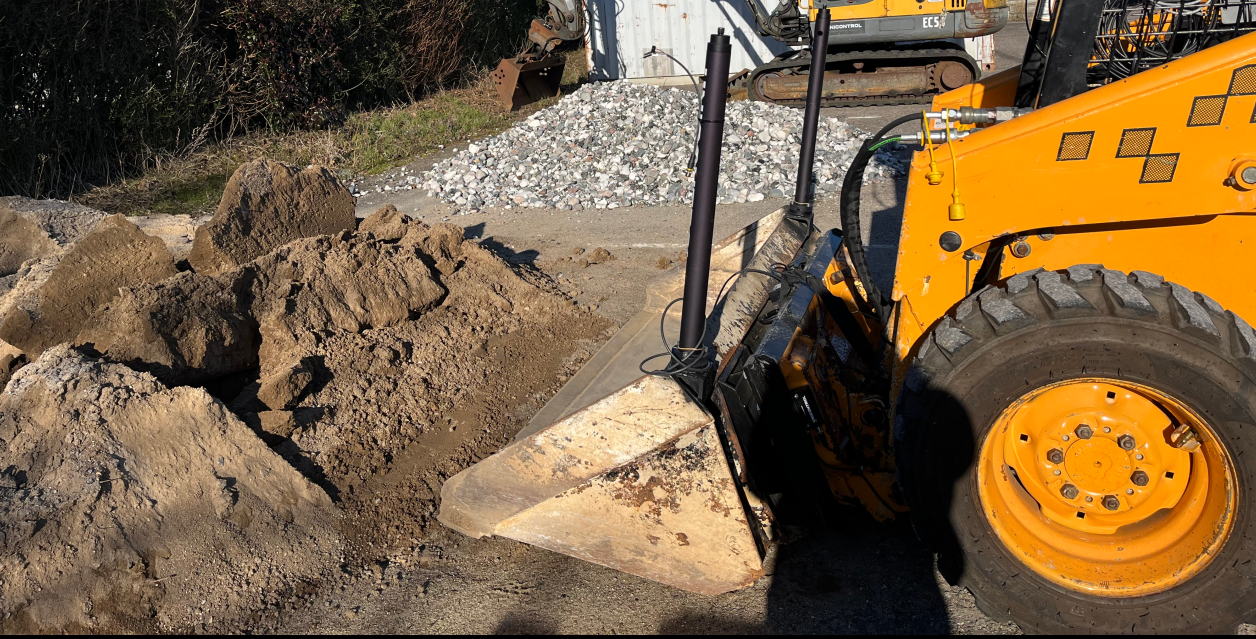}
    \caption{Experiment site with MUSTANG 2024 and wet sand and stone pile. }
    \label{fig:exp_cite}
\end{figure}

The reward convergence results learning the bucket loading task are shown in Fig.~\ref{fig:aem_exp3}.
The trained actor is deployed to a real machine: MUSTANG 2040, with wet sand and stone pile fields. The experiment site is shown in Fig.~\ref{fig:exp_cite}.

In the experiments, the inputs $[q_1, q_2]$ are measured in radians with Inertial Measurement Units (IMUs) mounted on the boom and bucket. $[L_a]$ denotes the forwarding distance of the loader, determined using GNSS antennas mounted on the machine. $[\hat{\tau}_e]$ is computed based on the pressure sensor readings obtained from both sides of the hydraulic pistons in the boom and bucket hydraulic pump. All the sensors operate at an update rate of $10 Hz$.
The output $[q_{d1},q_{d2},\tau_d]$ are from the deployed NNs, while the loader's forwarding motion is manually controlled by an operator at a random speed. The operator halts the forward motion upon noticing the boom's lift.

Firstly, a two-sided admittance controller with both position and torque reference is tested. However, due to the high compaction nature of wet sand and stone pile, the downward curl of the bucket causes dramatically large normal forces, the admittance controller fails to track $\tau_d$, thus leading the boom and bucket to vibrate during penetrating and unstable outputs from deployed actor network. These unstable NNs outputs could result from a state distribution shift, caused by the large normal forces during interacting with compacted material. In the simulation environment, such compaction effect is not accurately modeled, as the material pile is simulated using discrete particles that lack adhesive or cohesive properties. A penalty for causing such unsafe behavior should be considered in the future reward design. For the sake of safety and stable performance, only a one-sided admittance controller is tested in the following experiments with position reference $[q_{d1},q_{d2}]$ and a $\tau_{sat} = 800 $ N.m to prevent the bucket from getting stuck. 

\begin{figure}[htp]
    \centering    \includegraphics[width=8.0cm]{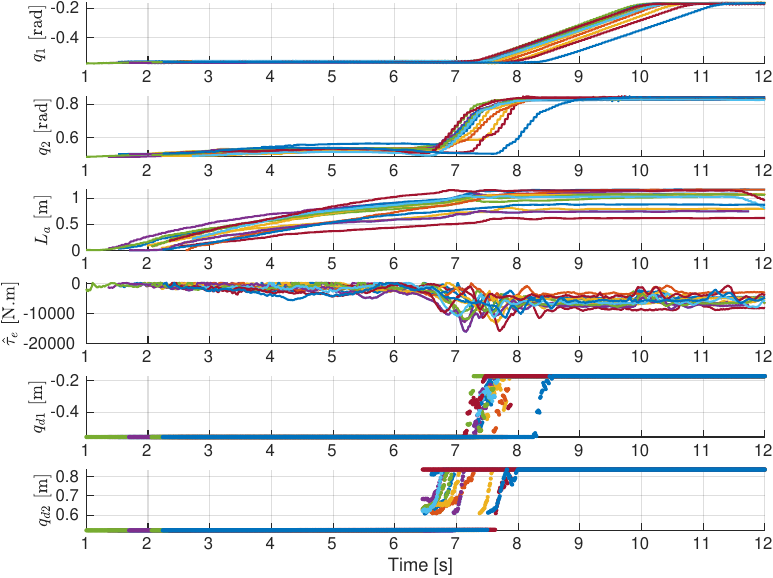}
    \caption{Sim2Real results with 15 times loadings of stones with different pile geometries. }
    \label{fig:exp}
\end{figure}

To evaluate the policy, 25 experiments are carried out, involving 10 trials for loading wet sand and 15 trials for loading stone. Sim-to-real results for loading stones are presented in Fig.~\ref{fig:exp}. Despite changing environments, including pile geometries, material types, and forwarding velocities, all the experiments successfully loaded and lifted the materials. The average bucket fill rates for loading sand and stone in the experiments are given in Table.~\ref{tab:Fill-rate}. To compare the sim2real performance in terms of bucket fill rate, the bucket fill rates in simulation are also recorded and averaged over 5 episodes. The bucket fill rate differences between simulation and experiments may stem from environmental uncertainties present in real-world conditions, such as the irregular pile shapes.
\begin{table}[h!]
  \centering
\begin{tabular}{ccccc}
  \hline
  Materials & simulation & experiment \\\hline
  Sand & 93.71\% & 85.81\%\\
  Stone & 90.01\% & 78.77 \% \\
  \hline
\end{tabular}
\caption{Average bucket fill rate in simulation and experiments. 
\label{tab:Fill-rate}}
\end{table}

\label{application}

% \section{Results}
% \input{results}

% \section{Experiment results}
% \input{Experiment results}

\section{Conclusion}
This paper proposes and implements an exploration-efficient \IBRLn{} framework to reduce the need for extensive interaction when applying off-policy DRL to real-world robotic tasks. The designed experiments empirically validate the effectiveness of our framework in mitigating bootstrapping errors and addressing convergence to suboptimal policies, ultimately reducing the exploration required to attain high-performing policies compared to IBRL.
Furthermore, we demonstrated the implementation details for using the DRLR framework on a real industrial robotics task, wheel loader bucket loading. The sim2real results validate the successful deployment of the considered framework, demonstrating its potential for application to complex robotic tasks.

In future work, one could improve the action selection module by selecting $s'_{t+1}$ by finding the states closest to $s_{t+1}$ within $\mathcal{D}$, by employing Euclidean or Mahalanobis distance, thereby facilitating more precise comparisons between neighboring state–action pairs. To better demonstrate the advantages of DRLR, it is necessary to compare it against established offline-to-online DRL baselines that explicitly addressed bootstrapping errors, such as CAL-QL, RLPD, and WSRL \cite{nakamoto2023cal, ball2023efficient, zhou2024efficient}.

Moreover, one could also consider using Deep Ensembles to quantify the uncertainties in the demonstrations and utilize these uncertainties as priors for SAC entropy. Integrating the concepts of Active Learning and Uncertainty-aware RL into the proposed framework could further improve the exploration efficiency.

\bibliography{references}
\bibliographystyle{ieeetr}

\section{Appendices}
\subsection{Experiments configurations}
\label{Sec.config}
\begin{table}[ht!]
\centering
\setlength{\tabcolsep}{4pt} % Column spacing
\renewcommand{\arraystretch}{1.1} % Row spacing
\scriptsize % Reduce font size to fit
\begin{tabular}{lcccc}
\hline
Configuration & \multicolumn{2}{c}{IBRL} & \multicolumn{2}{c}{\IBRLn{}} \\
& OpenDrawer & BucketLoad & OpenDrawer & BucketLoad \\
\hline
Learning rate & 3e-4 & 3e-4 & 3e-4 & 3e-4 \\
Batch size &  128 & 128 &  128 & 128 \\
Discount factor & 0.99 & 0.99 & 0.99 & 0.99 \\
Exploration noise Std. & 0.1 & 0.1 & -- & -- \\
Initial entropy & -- & -- & 0.1 & 0.01 \\
Learn entropy & -- & -- & True & False \\
Smooth noise Std. & 0.1 & 0.1 & -- & -- \\
Smooth noise clip & 0.5 & 0.5 & -- & -- \\
Dropout rate & 0.1 & 0.1 & -- & -- \\
Ensemble size of RED-Q & 5 & 5 & -- & -- \\
UTD & 5 & 5 & 1 & 1 \\
Replay buffer size & 300k & 200k & 300k & 200k \\
\hline
\end{tabular}
\caption{Configuration of IBRL and \IBRLn{} across two tasks. The code for replicate experiments $1 \sim 7$ for DRLR and IBRL are available at \url{https://github.com/impala-shen/DRLR}. 
Our RL methods are developed using the RL library: skrl \cite{serrano2023skrl}.}
\label{tab:config}
\end{table}

\subsection{Additional comparisons between IBRL and DRLR.}

To better understand the differences between IBRL and DRLR, additional comparisons including, 1) the Ref policy selection probabilities; 2) the bias of Q-return; 3) Mahalanobis Distance between sampled states to the expert states, are recorded for IBRL and DRLR. The additional comparisons are conducted on the new simulation platform: IsaacLab \cite{mittal2025isaaclab}, using a task called $FrankaCabinet$. Since IsaacGym is now deprecated, all experiments were migrated accordingly. The task is executed with $128$ parallel environments and trained over $5$ random seeds ($42–46$). Expert demonstrations are generated using a trained PPO policy. Although the current performance on this task is still suboptimal, future work will be done to improve the results. 

\begin{figure}[htb]
  \centering
  \begin{subfigure}[b]{0.45\columnwidth}
    \centering
    \includegraphics[width=\linewidth]{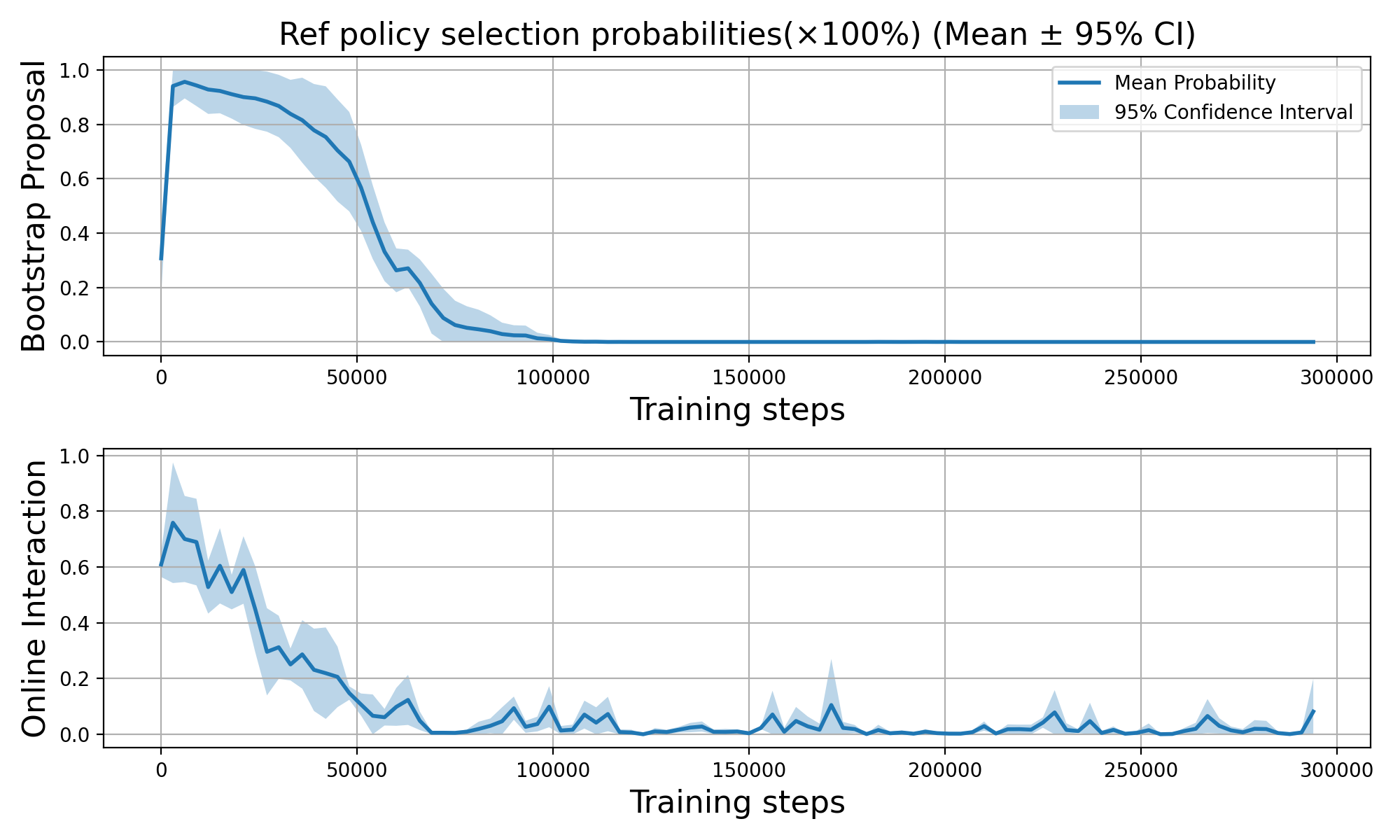}
    \caption{Visualization of the Ref policy selection probabilities in DRLR.}
    \label{fig:exp8_1}
  \end{subfigure}
  \hfill
  \begin{subfigure}[b]{0.45\columnwidth}
    \centering
    \includegraphics[width=\linewidth]{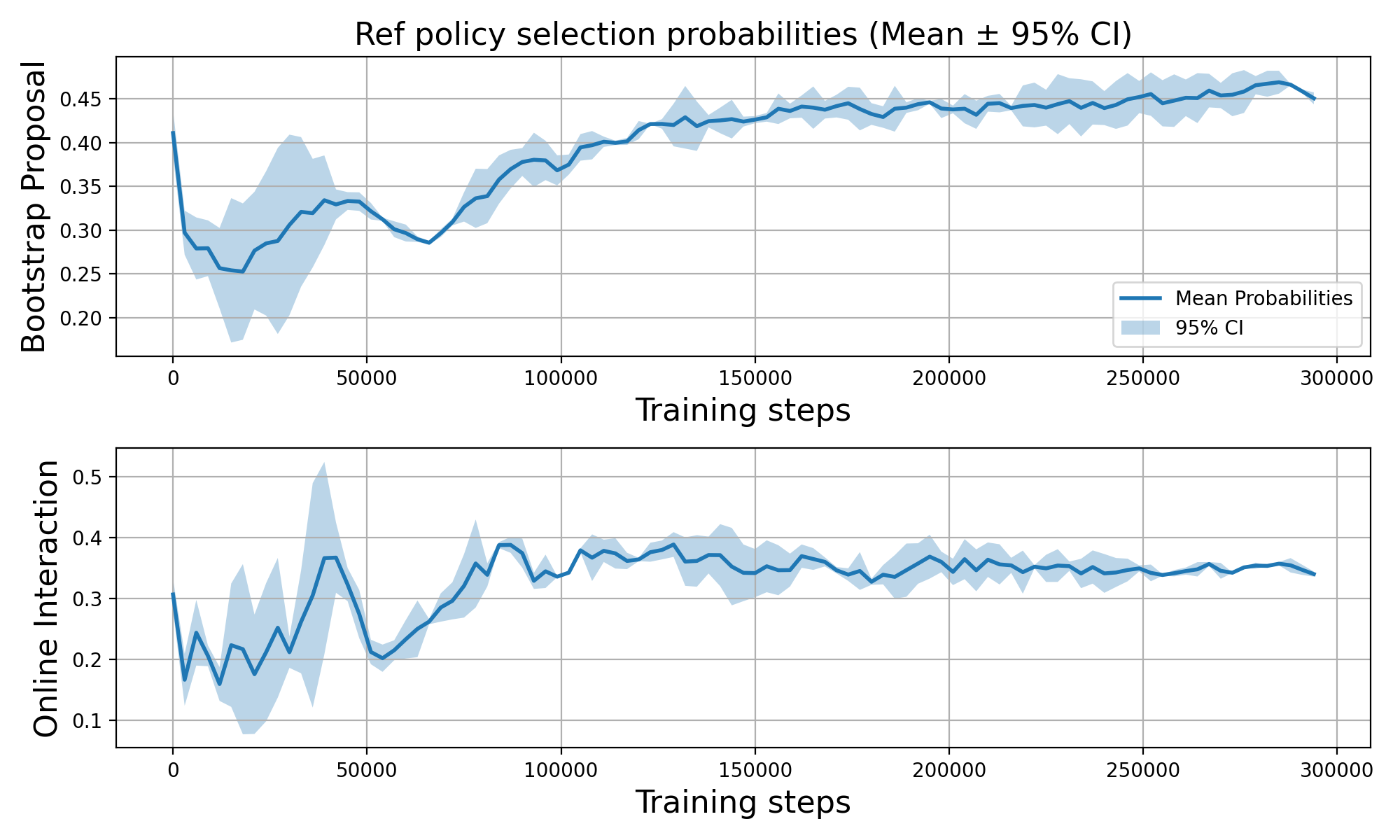}
    \caption{Visualization of the Ref policy selection probabilities in IBRL.}
    \label{fig:exp8_2}
  \end{subfigure}
  \caption{Comparison of the Ref policy selection probabilities in DRLR and IBRL.}
  \label{fig:exp8}
\end{figure}

Figure \ref{fig:exp8} presents the Ref policy selection probabilities for DRLR and soft IBRL with temperature $\beta = 1$. As shown in Fig. \ref{fig:exp8_1}, DRLR selects the reference policy aggressively during the first $5\times10^{4}$ training steps, after which the selection probability gradually decrease to zero. This behavior aligns with the proposed action-selection module: DRLR leverages the reference policy early to obtain high-reward samples quickly, and once the RL policy becomes competent, it quickly takes over, eliminating the need to rely on the reference policy.
In contrast, IBRL exhibits continuously increasing Ref policy selection throughout training, including near convergence. This trend indicates that IBRL remains dependent on the reference policies.

\begin{figure}[htb]
    \centering
\includegraphics[width=0.8\linewidth]{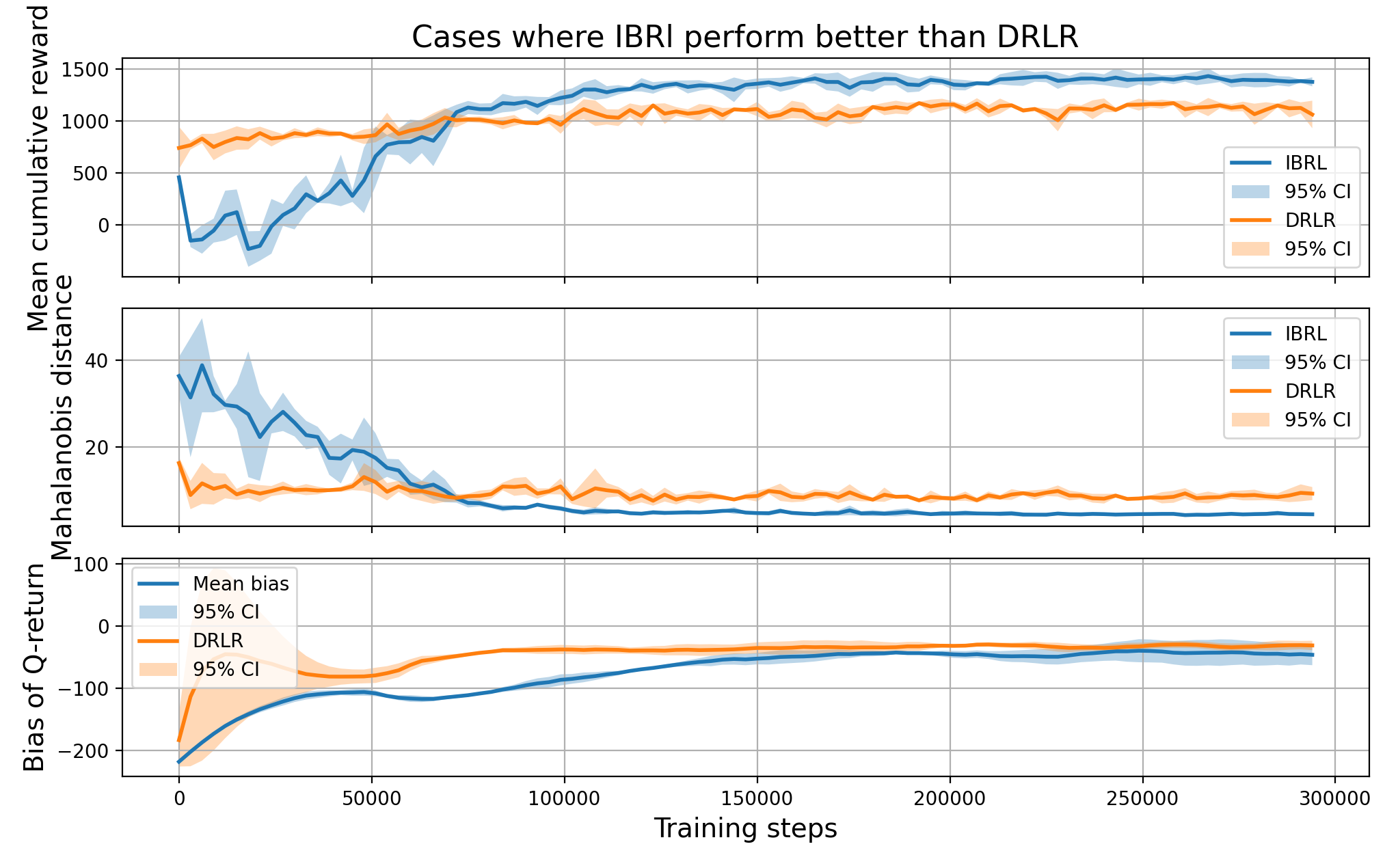}
  \caption{Cases where IBRL can obtain better reward convergence compared to DRLR because of its ability to obtain strong Ref policy. }
  \label{fig:exp9}
\end{figure}

However, because of the dependence of IBRL on the Ref policy, there are cases where IBRL can obtain better reward convergence compared to DRLR. This occurs when the Ref policy is fairly strong to accomplish the task and when the bootstrapping error during training is small.
To visualize these cases, 1) the Mahalanobis distance between sampled states and expert states, reflecting the state distribution shift, and 2) the bias of Q-return, are plot in Fig. \ref{fig:exp9}. In Fig. \ref{fig:exp9}, although DRLR exhibits a smaller Q-return bias (bottom plot), IBRL achieves better reward convergence (top plot) and a lower state distribution shift (middle plot). The lower state distribution shift in IBRL indicates that the optimized policy is close to the Ref policy.
While in DRLR, the mean Q-estimation of the RL policy initially catches up quickly with that of the Ref policy due to the high Ref policy selection rate at the beginning. However, once the RL policy rapidly takes over the learning process, it struggles to explore state–action pairs that could yield higher rewards, and converging to sub-optimal performance. Addressing this limitation in DRLR requires more effective RL online exploration strategies and more precise comparisons between neighboring state–action pairs.

\subsection{Admittance controller}
\label{Sec.appen_ad}
To control the wheel loader with an admittance controller, the wheel loader dynamics are modeled based on the Euler-Lagrange modeling:
\begin{equation}
    \begin{aligned}
    M(q_i)\ddot{q_i}+n(q_i,\dot{q_i}) = \tau_i + \tau_e,
    \end{aligned}
    \label{eqa:dynamics-d}
\end{equation}
Where
\begin{equation}
    \begin{aligned}
    &n(q_i,\dot{q_i}) = C(q_i,\dot{q_i})\dot{q_i}+\tau_{f}(\dot{q_i})+g(q_i) 
    \end{aligned}
    \label{eqa:dynamics-n}
\end{equation}
where $q_i, \dot{q}_i, \ddot{q}_i$ are position, velocity and acceleration of the joint, and the index $i = 1, 2$ is short for boom and bucket joint respectively. 
The non-linear effects, e.g. dead-zones caused by the hydraulic actuators are modeled as friction, $\tau_{f1}$ and $\tau_{f2}$ are torques caused by coulomb friction and viscous friction. $\tau_e$ is the external torque caused by interacting with the environment, it is estimated by a Sliding-mode Momentum Observer (MOB) proposed in \cite{shen2024safe}. 
The actuation torque $\tau_i$ can be obtained by the actuation force $F_{1}, F_{2}$ with the known hydraulic kinematics. $F_{i}$ is obtained based on \cite{Yu2023}:
\begin{align}
F_{i}=p_{base}A_{base}-p_{rod}A_{rod}
\end{align}
where the $p_{rod}, p_{base}$ are the pressure measurements from the pressure sensors installed on each side of the boom hydraulic cylinder. $A_{rod}, A_{base}$ are the approximate areas of the rod and base side of the cylinder.

\color{black}
\begin{figure}[htp]
    \centering
    \includegraphics[width=8.0cm]{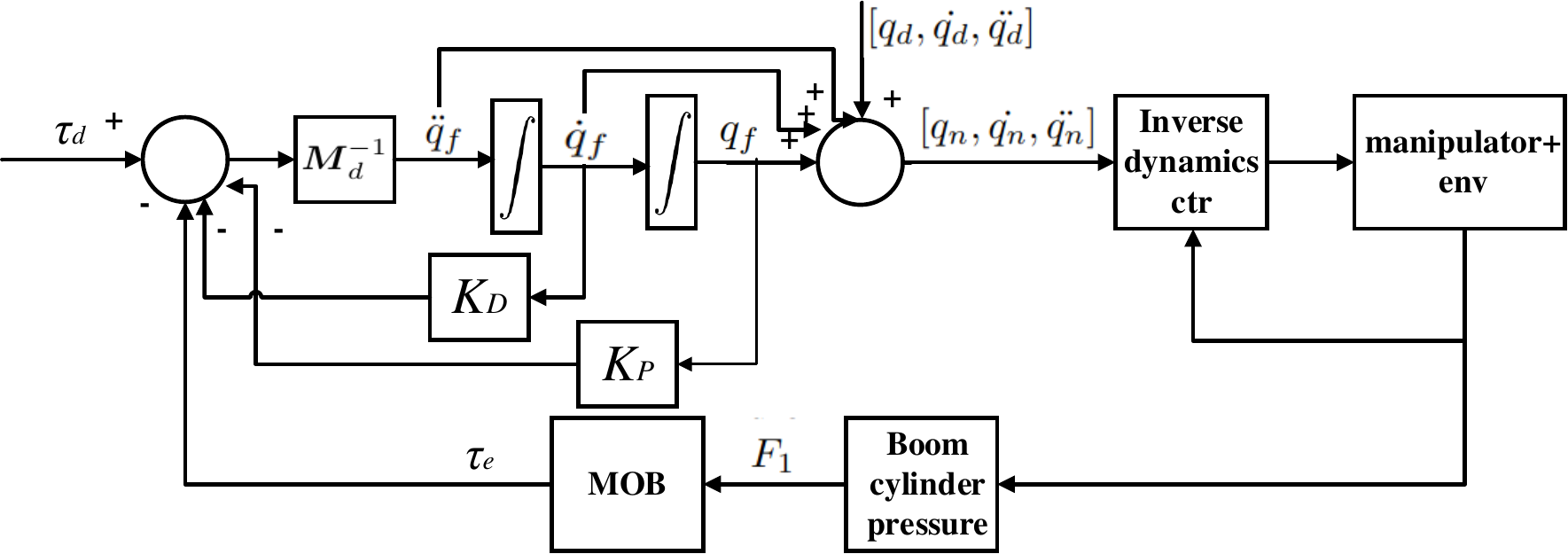}
    \caption{Proposed admittance controller.}
    \label{fig:adm}
\end{figure}
The admittance controller starts from the measurements of torque difference, a mechanical admittance is used to motion variables from torque difference.
The mechanical admittance law is given:
\begin{equation}
    \tau_d - \tau_e = M_d\ddot{q}_{f}+K_{Dt}\dot{q}_f +K_{P} q_f.
\end{equation}

\subsubsection{Two-sided}
According to \cite{Dobson2017}, a two-sided admittance control has the best loading efficiency compared to a manual operator. A two-sided admittance controller is designed:
\begin{equation}
    \ddot{q}_{f} =\left\{\begin{array}{ll}
    -M_{d}^{-1}((\tau_{sat} - \tau_e) - K_{D}\dot{q}_f -K_{P} q_f),& \hat{\tau}_{e} > \tau_{sat} \\
    M_{d}^{-1}((\tau_{d} - \tau_e) - K_{D}\dot{q}_f-K_{P} q_f), & else
    \end{array}\right.
\end{equation}
where $\tau_{sat}$ is to prevent the bucket's downward curl from lifting the wheel loader or causing dramatically large normal force. $\tau_{d}$ is loading reference torque, which is output by RL.

\subsubsection{One-sided}
To prevent the bucket's downward curl from lifting the wheel loader or causing dramatically large normal force, a one-sided admittance controller is also designed:
\begin{equation}
    \ddot{q}_{f} =\left\{\begin{array}{ll}
    -M_{d}^{-1}((\tau_{sat} - \tau_e) - K_{D}\dot{q}_f-K_{P} q_f) ,& \hat{\tau}_{e} > \tau_{sat} \\
    0, & else
    \end{array}\right.
\end{equation}

\end{document}